\documentclass[jair,twoside,11pt]{article}
\usepackage{jair, rawfonts}

\usepackage{theapa}
\usepackage[implicit=false]{hyperref}

\usepackage[utf8]{inputenc} 
\usepackage[T1]{fontenc}    
\usepackage{url}            
\usepackage{booktabs}       
\usepackage{amsfonts}       
\usepackage{nicefrac}       
\usepackage{microtype}      

\usepackage[colorinlistoftodos]{todonotes}

\usepackage{graphicx} 
\usepackage{float}
\usepackage{caption}
\usepackage{subcaption}
\usepackage{sidecap}
\usepackage{tabularx}
\usepackage{pbox}
\usepackage{csquotes}
\usepackage{booktabs}
\usepackage{xcolor}
\usepackage{wrapfig}
\usepackage{xcolor,colortbl}

\newcommand*\citep{\cite}
\usepackage{mathtools}
\usepackage{stackengine}

\usepackage{caption}

\usepackage{algorithm}
\usepackage{algorithm,algorithmicx}
\usepackage[noend]{algpseudocode}

\usepackage{enumitem}

\usepackage{amsmath, amssymb}

\usepackage{amsthm}
\usepackage{thmtools,thm-restate}
\usepackage{hyphenat}

\newcommand{\defeq}{:=}
\DeclareMathOperator*{\argmax}{arg\,max}

\jairheading{80}{2024}{441-473}{07/2023}{06/2024}
\ShortHeadings{Mitigating Value Hallucination in Dyna Planning via Multistep Predecessor Models}
{Aminmansour, Jafferjee, Imani, Talvitie, Bowling, \& White}
\firstpageno{441}

\begin{document}

\title{Mitigating Value Hallucination in Dyna-Style Planning via
Multistep Predecessor Models}

\author{\name Farzane Aminmansour \email aminmans@ualberta.ca\\
        \name Taher Jafferjee \email jafferje@ualberta.ca \\
       \name Ehsan Imani \email imani@ualberta.ca \\
       \addr   Dept of Computing Science \& the Alberta Machine Intelligence Inst\\
        University of Alberta, Canada 
       \AND
       \name Erin J. Talvitie \email erin@cs.hmc.edu  \\
       \addr Dept of Computer Science,\\
       Harvey Mudd College, USA
       \AND
       \name Michael Bowling \email mbowling@ualberta.ca \\
         \name Martha White \email whitem@ualberta.ca \\
       \addr Dept of Computing Science \& Amii \\
              University of Alberta, Canada}


\maketitle

\begin{abstract}
Dyna-style reinforcement learning (RL) agents improve sample efficiency over model-free RL agents by updating the value function with simulated experience generated by an environment model. However, it is often difficult to learn accurate models of environment dynamics, and even small errors may result in failure of Dyna agents. In this paper, we highlight that one potential cause of that failure is bootstrapping off of the values of simulated states, and introduce a new Dyna algorithm to avoid this failure. We discuss a design space of Dyna algorithms, based on using successor or predecessor models---simulating forwards or backwards---and using one-step or multi-step updates. Three of the variants have been explored, but surprisingly the fourth variant has not: using predecessor models with multi-step updates. We present the \emph{Hallucinated Value Hypothesis} (HVH): updating the values of real states towards values of simulated states can result in misleading action values which adversely affect the control policy. We discuss and evaluate all four variants of Dyna amongst which three update real states toward simulated states --- so potentially toward hallucinated values --- and our proposed approach, which does not. The experimental results provide evidence for the HVH, and suggest that using predecessor models with multi-step updates is a promising direction toward developing Dyna algorithms that are more robust to model error.
\end{abstract}

\section{Introduction}\label{Section:Introduction}

Sample efficient algorithms for reinforcement learning (RL) are critical for the use of RL in most realistic settings. RL agents have been developed that can achieve impressive performance in simulated problems, including agents that can play arcade games \cite{Bellemare} at human level performance \cite{Mnih} and robotic hands with human-like dexterity \cite{Andrychowicz}.
These successes have come at a cost: large data requirements. For example, DQN requires $200$ million samples to learn control policies for
games. Such sample requirements make RL an unviable solution for many
realistic settings, and a key challenge is developing sample efficient
RL methods.

One promising path to address this challenge is Dyna \cite{Sutton1990}: an architecture where real experience is supplemented with simulated experience.
In Dyna, agents use a model of their environment to simulate experience. By updating their value functions with this data, in a process known as \emph{planning}, agents may learn improved control policies using less real data than model-free RL agents.

Early work on Dyna considered environments with deterministic dynamics and finite state spaces \citep{Sutton1990,Sutton1991}. In such settings, learning perfectly accurate environment models is possible: store a table of states and actions and record the next observed states for a given state and action. Generally, however, learning perfect environment models is difficult, if not impossible. Moreover, emerging evidence suggests that even small model errors can result in large errors in the value function \cite{Van}. 

Most work investigating how to minimise model error has primarily considered errors due to model rollouts. 
In rollouts, a model's predictions are iteratively fed back to itself to generate a simulated trajectory. Even small inaccuracies in predictions can compound and render a trajectory useless for planning as the final predictions bear little resemblance to any plausible future as shown by \citeA{Talvitie2014}, \citeA{Talvitie2017}. To address this, Talvitie proposed Hallucinated Replay: training a model on its own imperfect predictions to minimise compounding error. Related approaches have been considered in time series modeling \citep{venkatraman2015improving}. Other work reduced error due to iteration by learning a model from a new representation of states in a latent space rather than learning a model from original observations coming from the environment \cite{Ke,hafner2023mastering,hafner2022mastering,serban2020bottleneck}, with some further restricting the use of the model only for policy optimization \cite{amos2021model,tomar2021model}. 

Improving model accuracy, however, is unlikely to resolve failure in Dyna. For example, \citeA{Holland} showed that even with a learning curriculum to minimise compounding error, imperfections in model predictions eventually overwhelm any signal in simulated trajectories and value function updates with this data is detrimental to learning. \citeA{Van} showed that updating action-values with an inaccurate model could result in poor performance; to avoid these failures, they suggested that simulating transitions forward should only be used for decision-time planning, rather than updating the action-value. A direction being actively pursued is to learn models based on value error, rather than prediction error on the next state \cite{farahmand2017value,farahmand2018iterative,ayoub2020model,grimm2020value}, or other objectives better aligned with the goals of planning \cite{lambert2020objective,abachi2020policy}. 

Another strategy to deal with model error has been to downweight or filter potentially erroneous simulated data. One direction has been to limit the use of the model and only use it when in areas of high uncertainty \cite{Kalweit}, such as in early learning. 
Another idea is to correct the effect of small model errors by using samples from the true model \cite{rakhsha2022operator}.
Other work tried to limit errors introduced into the bootstrap targets for learning the value function, by both (a) strictly limiting the iteration horizon for the model and (b) using an ensemble of bootstrapped targets by averaging all $n$-step bootstrapped targets up to the iteration horizon \cite{feinberg2018model}. Follow-up work improved the algorithm by using residual gradient algorithms, which are more stable under distribution mismatch \cite{zhang2019deep}. The ensemble approach, however, did not directly account for model error. Later work used model-error to further avoid high-error bootstrap targets, either by dynamically truncating rollouts \cite{Buckman} or down-weighting uncertain updates \cite{abbas2020selective}. In general, knowing how sensitive Model-Based Reinforcement Learning (MBRL) is to model error \cite{Van}, much more work needs to be done to understand the sources of these catastrophic failures and to develop methods that are able to prevent them.

This paper identifies and examines a source of MBRL failure that has not been previously articulated---Hallucinated Values---and proposes a novel Dyna-style algorithm to mitigate this error, called Multi-step Predecessor Dyna. The issue we highlight is that simulated states can be unreachable, or rarely visited, resulting in arbitrary or undertrained values in those states. The agent is misled when it bootstraps off of these hallucinated values. The proposed algorithm is simple: using multi-step bootstrapped returns and a predecessor model to simulate transitions backwards from observed states. This approach ensures that the agent always bootstraps off of values in real states, that were updated when observed. This contrasts existing Dyna variants, that either use one-step updates or use multi-step updates but with successor models---those that simulate forward. 

We introduce an environment to test the hypothesis and show that previous variants of Dyna fail when the model is imperfect whereas our algorithm does not. We further test the algorithms on three classic benchmark environments and find even in these environments the same behavior persists. This success in environments not explicitly designed to suffer from hallucinated values suggests both that these phenomena may be prevalent---even if hard to directly measure---and that Multi-step Predecessor Dyna may provide additional benefits by improving credit assignment with multi-step updates. 

This paper is organised as follows. In
Section \ref{Section:Background} we formalize the learning problem and overview model-free and model-based RL methods. Section \ref{Section:ADynaDesignSpaceOfDynaAlgorithms} provides a general framework for Dyna-style algorithms in which four variants of planning approaches are discussed. We propose a new method called Multi-step Predecessor Dyna. 
Section \ref{Section:Hypothesis} introduces the Hallucinated Value Hypothesis (HVH), and discusses how variants of Dyna can suffer from it. In Section \ref{Section:Experiments}, we demonstrate the robustness of our approach with respect to HVH by evaluating it on a specifically designed environment, called BorderWorld, and three other benchmarks. These benchmarks are selected both (a) to understand the prevalence of the HVH, in standard environments where it is not obvious it will be a problem and (b) to more generally investigate the utility of our multi-step predecessor approach in classic benchmark problems. Section \ref{Section:beta} analyzes the robustness to model iteration, in all of the Dyna variants.
Finally, Section \ref{Section:Conclusion} wraps up the paper by providing a conclusion and a discussion on the limitations and future directions.  

\section{Problem Formulation}\label{Section:Background}

We formalized the control problem using a {\em Markov Decision Process} (MDP). MDPs are defined by the tuple $(\mathcal{S}, \mathcal{A}, r, p, \gamma)$. $\mathcal{S}$ is a set of states, $\mathcal{A}$ is a set of actions, $r: \mathcal{S} \times \mathcal{A} \times \mathcal{S} \mapsto \mathbb{R}$ is a reward function defining the reward received upon taking a given action in a given state, $p: \mathcal{S} \times \mathcal{A} \times \mathcal{S} \mapsto [0, 1]$ is a function defining the transition dynamics of the MDP, and $\gamma \in [0, 1)$ is a discount parameter influencing how far-sighted or myopic the agent is. At time $t$, the agent is in a state $s_t \in \mathcal{S}$, takes action $a_t \in \mathcal{A}$,  transitions to a new state $s_{t+1} \in \mathcal{S}$ with probability $p(s_t, a_t, s_{t+1})$ and receives reward $r_{t+1} = r(s_t, a_t, s_{t+1})$. 

The agent selects actions according to a policy $\pi:\mathcal{S}\times\mathcal{A} \mapsto [0, 1]$, where $\pi(\cdot | s)$ is a distribution over actions in state $s$. The agent's goal is to learn a policy which maximizes its expected return, represented by the action-value function
\begin{equation*}
    Q^{\pi}(s,a) \defeq \mathbb{E}\left[\sum_{k = 0}^{\infty} \gamma^{k} R_{t+k+1} | S_t = s, A_t = a\right] = \mathbb{E}\left[R_{t+1} + \gamma V^{\pi}(S_{t+1}) | S_t = s, A_t = a\right]
\end{equation*} 
where uppercase letters, such as $R_{t+1}$, indicate random variables and probabilities are according to $p$ and $\pi$. The action-value $Q^{\pi}(s,a)$ is the expected return from state $s$, after taking action $a$ and following policy $\pi$ thereafter. 

Q-learning \cite{Watkins} is an approach to learn the optimal policy, by iteratively updating towards $Q^*$ the action-values for the optimal policy. The agent estimates the action-values as a parameterized function, $Q$, with parameters $w$,
using Q-learning update
\begin{align*}
    \delta &\gets r_{t+1} + \gamma \max_a Q(s_{t+1}, a) - Q(s_{t}, a_{t})\\
    w &\gets w + \alpha \delta \nabla Q(s_{t}, a_{t})
\end{align*} 
To keep notation simple thorough-out, we do not explicitly show $Q$ as a function of $w$, and assume that the gradient $\nabla$ is always with respect to $w$.
 Q-learning is a {\em model-free} algorithm, i.e., it updates $Q$ using only from data gathered by interacting with the environment.

The Dyna framework \cite{Sutton1990} is a {\em model-based approach}, that updates the action-values with both real experience and transitions sampled from a learned model. 
In Dyna-Q, summarized in Algorithm \ref{Algorithm:Original-Dyna-Q}, an agent first does a Q-learning update on real experience and updates its model with this new data. Then it performs planning updates in which it 1) draws a state $s_i$ from its previous experience; 2) selects action $\hat{a}_i$ to perform in $s_i$ using its current policy; 3) uses its model to generate next state $\hat{s}_{i+1}$ and reward $\hat{r}_{i}$ (hat indicates it is a simulated variable); and 4) performs the Q-learning update on $(s_i, \hat{a}_i, \hat{r}_i, \hat{s}_{i+1})$.  %
These extra updates should help improve sample efficiency. The idea behind Dyna is simple and elegant: simulated data can be treated like real experience, allow learning and planning to be interleaved. Since the introduction of the original Dyna-Q algorithm \cite{Sutton1990}, many variants of Dyna have been proposed. In Section \ref{Section:ADynaDesignSpaceOfDynaAlgorithms}, we discuss a general design space of Dyna-Q algorithms that incorporates many of these variants.

\begin{algorithm}[t]
	\caption{Original Dyna-Q}
	\label{Algorithm:Original-Dyna-Q}
	\begin{algorithmic}[1]
		\State \textbf{Input} Subroutines \texttt{\textbf{update\_model}}, 
		\State Initialise action-value function $Q$, and environment model  $\mathcal{M}$
		\State Set learning rate $\alpha$
		\While{Agent Interacting with Environment}
		\State Observe state $s$, select action $a$ using policy $\pi$
		\State Observe reward $r$ and next state $s'$
		\State $\delta \gets r + \gamma \max_{a'} Q(s', a') - Q(s, a)$
		\State $Q(s, a) \gets Q(s, a) + \alpha \delta $
		\State \texttt{\textbf{update\_model}}($\mathcal{M}, (s, a, r, s')$)
               
		\For{$\text{planning step} = 1 \text{ to } N$}
		\State $s \gets \text{random previously observed state}$
		\State $a \gets \text{random action previously taken in s}$
		\State $(s', r)  \gets \mathcal{M}  (s, a)$
		\State $\delta \gets r + \gamma \max_{a'} Q(s', a') - Q(s, a)$
		\State $Q(s, a) \gets Q(s, a) + \alpha \delta $
		\EndFor
		\EndWhile
	\end{algorithmic}
\end{algorithm}

\section{A Design Space of Dyna Algorithms}\label{Section:ADynaDesignSpaceOfDynaAlgorithms}
Dyna is a flexible framework which admits a variety of implementations. In this work we focus on two design choices that provide four Dyna variants: 1) simulating forward or backward in time and 2) using one-step or multi-step return targets in the action-value update during planning. We explain these choices below but otherwise make standard choices for Dyna: updating with real experience, using prioritization, and simulating all actions during planning.

The four algorithms are variations of Algorithm \ref{Algorithm:Prioritised-Dyna}.
Real environment data is used to update the action-value functions and the environment model (lines 7-9). The real experience is added to a search-control queue $P$, with priority $|\delta |$ (line 10) if the priority is not smaller than a threshold $\rho$. The transitions in this queue are used to perform $N$ planning steps (lines 11 - 15). In each planning step the highest priority transition is popped off the queue (line 12), the model $\mathcal{M}$ is used to generate simulated transitions (line 13), and these transitions are used to update action-values (line 14). Finally, these simulated transitions are added to the queue with a priority equal to their TD error multiplied by $\beta^n$ (line 15) for $\beta \in [0, 1]$ and $n$ is the number of times the model has been iterated to obtain that trajectory. This insertion happens unless the computed priority is smaller than $\rho$

We introduce the decay rate $\beta$ to reduce priority for trajectories that have been iterated for many steps. This prevents highly iterated trajectories from being added to $P$ thereby mitigating compounding model error. This gradual decay avoids the need to specify a strict horizon for iteration, and uses the existing mechanism of priorities to implicitly control the horizon of model iteration. In Subsection \ref{Dynas} we discuss the implications of $\beta$ on the Dyna variants that we study.

This Dyna algorithm is standard except for $\beta$ and the fact that entire tuples are stored on the queue with an explicit parameter $n$. This storing of tuples is a necessary modification to consider multi-step updates with Dyna, as described in Section \ref{sec_step}. The different Dyna variants execute the planning update differently, given in Algorithm \ref{Algorithm:FQ1-Planning-TD-Update}; we explain these different updates in the next sections. 

\begin{algorithm}[t]
	\caption{Prioritised-Dyna with Multi-step Updates}
	\label{Algorithm:Prioritised-Dyna}
	\begin{algorithmic}[1]
		\State \textbf{Input} Subroutines \texttt{\textbf{update\_model}}, \texttt{\textbf{planning\_update}}  \Comment{Differ for variants} 
		\State Initialise action-value function $Q$, prioritised planning queue $P$, environment model $\mathcal{M}$, planning $mode$
		\State Set learning rate $\alpha$, priority threshold $\rho$, decay $\beta \in [0,1]$ (default: $\rho  = 0.01$, $\beta = 0.5$) 
		\While{Agent Interacting with Environment}
		\State Observe state $s$, select action $a$ using policy $\pi$
		\State Observe reward $r$ and next state $s'$
		\State $\delta \gets r + \gamma \max_{a'} Q(s', a') - Q(s, a)$
		\State $Q(s, a) \gets Q(s, a) + \alpha \delta $
		\State \texttt{\textbf{update\_model}}($\mathcal{M}, (s, a, r, s')$)
                \If{$|\delta| \geq \rho$}
                \State Insert $((s, a, r, s'), n = 0)$ to $P$ with priority $|\delta|$
                \EndIf
		\For{$\text{planning step} = 1 \text{ to } N$} 
		\State $T', \delta_{{T^{'}}} =$ \texttt{\textbf{planning\_update}}($P$) \Comment{See Algorithm \ref{Algorithm:FQ1-Planning-TD-Update}}
                \If{$| \delta_{{T^{'}}} |\beta^n \geq \rho$}
                \State Insert ${T'}$ to $P$ with priority $|\delta_{{T^{'}}}|\beta^n$
                \EndIf

		\EndFor
		\EndWhile
	\end{algorithmic}
\end{algorithm}

\begin{algorithm}[t]
	\caption{Planning Update}
	\label{Algorithm:FQ1-Planning-TD-Update}
	\begin{algorithmic}[1]
		\State \textbf{Input} Priority queue $P$ to pop a trajectory $T_n = (s_{t}, a_{t}, r_{t+1}, ..., s_{t+n}, n)$
		\If{Successor}
		\If{One-Step}
            \State Pop tuple $T_n = (s_t, a_t, ..., s_{t+n}, n)$ from $P$ 
            
            \For{$a_{i}$ in $\mathcal{A}$}
            \State Simulate dynamics forward $\hat{s}_{t+n+1}, \hat{r}_{t+n} = \mathcal{M}(s_{t+n}, a_{i})$
		\State $T_{n+1} = (s_{t}, a_{t}, r_{t+1}, ..., s_{t+n}, a_{i}, \hat{r}_{t+n}, \hat{s}_{t+n+1}, n+1)$
		\State $\delta \gets \hat{r}_{t+n} + \gamma \max_{a'} Q(\hat{s}_{t+n+1}, a') - Q(s_{t+n}, a_{i})$
		\State $Q(s_{t+n}, a_{i}) \gets Q(s_{t+n}, a_{i}) + \alpha \delta \nabla Q(s_{t+n}, a_{i}) $
  		\EndFor

		\EndIf
		\If {Multi-Step}
            \State $T_n \texttt{, }expanding \gets \texttt{\textbf{Screening}}(P)$
            \For{$a_{i}$ in $\mathcal{A}$}
		\State Simulate dynamics forward $\hat{s}_{t+n+1}, \hat{r}_{t+n} = \mathcal{M}(s_{t+n}, a_{i})$
		\State $T_{n+1} = (s_{t}, a_{t}, r_{t+1}, ..., s_{t+n}, a_{i}, \hat{r}_{t+n}, \hat{s}_{t+n+1}, n+1)$
		\State $\delta  \gets \sum_{k=0}^{n} \gamma^k \hat{r}_{t+k} + \gamma^{n+1} \max_a Q(\hat{s}_{t+n+1},a) - Q({s}_{t}, {a}_{t})$
		\State $Q(s_{t}, a_{t}) \gets Q(s_{t}, a_{t}) + \alpha \delta \nabla Q(s_{t}, a_{t}) $
            \EndFor
            \If {$\texttt{NOT } expanding$}
            \State $\delta \gets 0$
            \EndIf
		\EndIf
		\EndIf
		\If{Predecessor}
		\If{One-Step}
            \State Pop tuple $T = (s_t, a_t, ..., s_{t+n}, n)$ from $P$ 

             \For{$a_{i}$ in $\mathcal{A}$}
            \State Simulate dynamics backward $\hat{s}_{t-1}, \hat{r}_{t-1} = \mathcal{M}(s_{t}, a_i)$
		\State $T_{n+1} = (\hat{s}_{t-1}, a_i, \hat{r}_{t-1}, s_t, ... s_{t+n}, n+1)$
		\State $\delta \gets \hat{r}_{t-1} + \gamma \max_{a} Q({s}_{t}, a_{t}) - Q(\hat{s}_{t-1}, a_{i})$
		\State $Q(\hat s_{t-1}, a_{i}) \gets Q(\hat s_{t-1}, a_{i}) + \alpha \delta \nabla Q(\hat s_{t-1}, a_{i})$
            \EndFor
		\EndIf
		\If {Multi-Step}
            \State $T_n \texttt{, }expanding \gets \texttt{\textbf{Screening}}(P)$
            \For{$a_{i}$ in $\mathcal{A}$}
            \State Simulate dynamics backward $\hat{s}_{t-1}, \hat{r}_{t-1} = \mathcal{M}(s_{t}, a_i)$
		\State $T_{n+1} = (\hat{s}_{t-1}, a_i, \hat{r}_{t-1}, s_t, ... s_{t+n}, n+1)$
		\State $\delta \gets  \sum_{k=-1}^{n} \gamma^{k+1} \hat{r}_{t+k} +  \gamma^{n} \max_a Q({s}_{t+n-1},a_{t+n-1}) - Q(\hat{s}_{t-1}, {a}_{i})$
		
		\State $Q(\hat{s}_{t-1}, a_{i}) \gets Q(\hat{s}_{t-1}, a_{i}) + \alpha\delta \nabla Q(\hat{s}_{t-1}, a_{i})$
		\EndFor
            \If {$\texttt{NOT } expanding$}
            \State $\delta \gets 0$
            \EndIf
		\EndIf
		\EndIf
  		\Return $T_{n+1}, \delta$
	\end{algorithmic}
\end{algorithm}

\subsection{Successor vs Predecessor Models}
The first design choice we explore is whether the model simulates environment dynamics forward in time or backward in time. A model that simulates environment dynamics forward in time is a {\em successor model}. From state $s_t$ given action $\hat{a}_t$, the successor model generates a successor state $\hat{s}_{t+1}$ and reward $\hat{r}_{t+1}$. Conversely, a \emph{predecessor model} simulates dynamics backward in time. From state $s_{t}$, given action $\hat{a}_{t-1}$, a predecessor state $\hat{s}_{t-1}$ and reward $\hat{r}_{t}$ are generated.

We can feed back a model's predictions to itself in order to generate subsequent predictions. This leads to an iteration process in which we generate trajectories that radiate forward or backward from a state. Suppose we use a successor model in Algorithm \ref{Algorithm:Prioritised-Dyna}, and during planning a tuple $(s_t, a_t, r_{t+1}, s_{t+1}, n = 0)$ is popped from $P$ ($n = 0$ indicates this trajectory has not been iterated). The model could be queried to yield the first iterated transition $(s_t, a_t, r_{t+1}, s_{t+1}, \hat{a}_{t+1}, \hat{r}_{t+2}, \hat{s}_{t+2}, n = 1)$ which could be added to $P$. This tuple may be popped in the next planning step and the model queried to yield a second iterated transition $(s_t, a_t, r_{t+1}, s_{t+1}, \hat{a}_{t+1}, \hat{r}_{t+2}, \hat{s}_{t+2}, \hat{a}_{t+2}, \hat{r}_{t+3}, \hat{s}_{t+3},  n=2)$. This process may be continued to generate tuples for all possible actions yielding trajectories that radiate from $s_t$.

We can also examine tuples generated when iterating a predecessor model. Again, suppose the first tuple is popped from the queue is $(s_t, a_t, r_{t+1}, s_{t+1}, n= 0)$. The predecessor model produces an iterated transition, $(\hat{s}_{t-1}, \hat{a}_{t-1}, \hat{r}_{t}, s_t, a_t, r_{t+1}, s_{t+1}, n= 1)$. This tuple could be added to the queue, popped and iterated to yield  $(\hat{s}_{t-2}, \hat{a}_{t-2}, \hat{r}_{t-1}, \hat{s}_{t-1}, \hat{a}_{t-1}, \hat{r}_{t}, {s}_{t}, {a}_{t}, r_{t+1}, s_{t+1}, n= 2)$ and so on. 

Iterating model predictions in this manner is typical of many Dyna algorithms. Some algorithms use successor models with iteration \citep{Gu} while others such as Prioritized Sweeping \citep{Moore,Peng} use a predecessor model to iterate predictions backwards in time. Since early work on prioritized sweeping, several papers have explored learning and using predecessor models \cite{Sutton2008,edwards2018forwardbackward,Pan,Van}. Two recent papers explicitly highlight that the one of the primary benefits of using a model-based approach, over simply using experience replay, is to facilitate iterating backward in time from high priority states \cite{Pan,Van}.

It is important to note that an open question remains about how to precisely define predecessor models. Given a transition $(s,a,r, s')$, it is straightforward to either use this data to update a successor model with input $(s,a)$ and output $(r, s')$ or to update a predecessor model with input $(s', a)$ and output $(s, r)$. Even with a changing policy, the target distribution for the successor model is stationary: it is $p(s', r | s, a)$ for a distribution model or $\mathbb{E}[(S', R) | S = s, A = a]$ for an expectation model. For the predecessor model, it is not clear what the target distribution is for a changing policy because the distribution over the predecessor states $s$ changes with the policy. Imagine a situation where both $s_1$ and $s_2$ lead to $s'$ when taking action $a$. If a policy visits state $s_1$ more frequently than $s_2$, then the predecessor model will predict $p(s_1 | s', a)$ higher than $p(s_2 | s', a)$. If we simply update on the data without any reweighting, this distribution is inherently tied to the policy. This distribution is inherently tied to the policy; if we simply update on the data without any reweighting. 

In this work, we partially sidestep this issue by gathering data for the model under a fixed policy. This strategy at least makes the target predecessor model stationary. It does not address the impact of the state visitation distribution of the policy on the utility of the predecessor model, nor if a different policy could have produced a more useful predecessor model. Such an investigation into learning predecessor models should be explored in a dedicated work on this open question and is outside the scope of this work. 

\subsection{One-Step vs Multi-Step Updates}\label{sec_step}
The second choice we examine is if the algorithm performs one-step updates or multi-step updates. A one-step update bootstraps immediately on the next state, whereas a multi-step update sums up multiple rewards until finally bootstrapping on a state multiple steps into the future.  

To see how this choice manifests in Dyna, consider a successor model starting with $(s_t, a_t, r_{t+1}, s_{t+1}, n= 0)$. From $s_{t+1}$ the successor model extends the trajectory to get $(s_t, a_t, r_{t+1}, s_{t+1}, \hat{a}_{t+1}, \hat{r}_{t+2}, \hat{s}_{t+2}, n = 1)$. The one-step update would use $\hat{s}_{t+2}$ to update $s_{t+1}$:
\begin{align*}
\delta &\gets \hat{r}_{t+2} + \gamma \max_a Q(\hat{s}_{t+2}, a) - Q(s_{t+1}, \hat{a}_{t+1})\\
Q(s_{t+1}, \hat{a}_{t+1}) &\gets Q(s_{t+1}, \hat{a}_{t+1}) + \alpha \delta \nabla Q(s_{t+1}, \hat{a}_{t+1})
\end{align*}
The multi-step update, however, would update $s_t$ to $\hat{s}_{t+2}$, with the discounted sum of rewards in-between
\begin{align*}
\delta &\gets r_{t+1} + \gamma \hat{r}_{t+2} + \gamma^2 \max_a Q(\hat{s}_{t+2},a) - Q(s_{t}, a_{t})\\
Q(s_t, a_t) &\gets Q(s_t, a_t) + \alpha \delta \nabla Q(s_t, a_t)
\end{align*}
Imagine this tuple is iterated again. The one-step update would update the value of $\hat{s}_{t+2}$ towards the value of $\hat{s}_{t+3}$ whereas the multi-step update would update the value of ${s}_{t}$ towards the value of $\hat{s}_{t+3}$. 
Both update approaches use the same generated trajectory, but use it differently. One-step TD performs all the one-step updates along the trajectory, updating the values for all the states in the
trajectory. Multi-step TD performs a one-step update, then a two-step update, then a three-step update and so on, all updating only state
$s_t$. 

This update, however, does not correct for the fact that the sequence of actions in the trajectory may not correspond to actions taken by the greedy policy under the current action-values.
This biased update is actually a standard strategy to use Q-learning with multi-step updates. An alternative is to truncate the multi-step update early, using only the part of the trajectory that is on-policy. The idea is to compare each action of a trajectory in the buffer with the greedy actions according to the current action-value function. We get a sub-chunk of a trajectory that has a chain of on-policy actions according to the current action-values. This is like incorporating importance sampling, where we re-weight the updates using a ratio between the target and behavior policies, because for the greedy actions under Q-learning, the importance ratios are 1 or 0. In the Dyna variants listed below, we highlight where this \emph{trajectory screening} choice is made, and leave it as a generic option. The detailed algorithms for different variants of screening approaches can be found in \ref{appendix A}. In our own experiments, we tested both, and generally found the truncated multi-step approach to be more effective, and that using the biased, uncorrected trajectories sometimes caused poor performance.

\begin{figure}
	\begin{center}
		\subfloat[One-Step Successor Dyna]{\includegraphics[width=0.35\columnwidth]{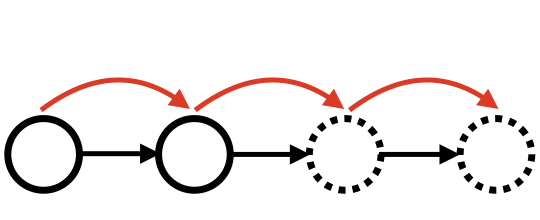}}
		\hspace{1em}
		\subfloat[Multi-Step Successor Dyna]{\includegraphics[width=0.35\columnwidth]{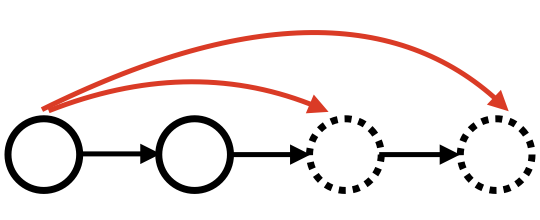}}\\
		\subfloat[One-Step Predecessor Dyna]{\includegraphics[width=0.35\columnwidth]{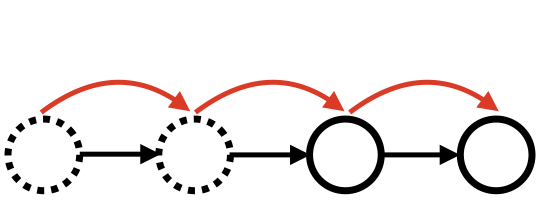}}
		\hspace{1em}
		\subfloat[Multi-Step Predecessor Dyna]{\includegraphics[width=0.35\columnwidth]{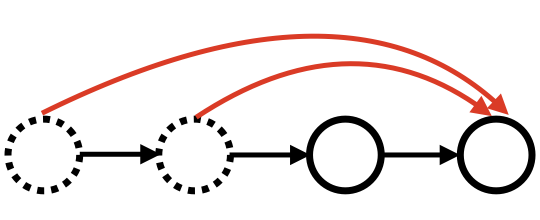}}
		\caption{A visual comparison of the planning updates in four Dyna algorithms. Circles and black arrows show a trajectory; solid circles are real states and dashed circles are simulated states. A red arrow means that the value of the originating state is updated towards the destination state. All algorithms except Multi-Step Predecessor Dyna allow updates towards simulated states, which we show in Section \ref{Section:Hypothesis} that these approaches can suffer from updating towards hallucinated values.
		}
		\label{Figure:Section4}
	\end{center}
\end{figure}

\subsection{The Four Dyna Variants}\label{Dynas}
The four variants are identical except in parts of the planning loop (Lines 11 - 15 of Algorithm \ref{Algorithm:Prioritised-Dyna}), so we focus our discussion below on the differences in planning. When describing each variant we also discuss the related work that used such an update. We summarize the literature in Table \ref{Table:Section4/Algo_Quadrants}. Moreover, Figure \ref{Figure:Section4} visually compares the four algorithms.  

\textbf{One-Step Successor} uses a successor model with one-step updates. Suppose a tuple $T_n = (s_t, a_t, r_{t+1}, s_{t+1}, n=0)$ is popped from $P$. We use the model $\mathcal{M}$ to extend the trajectory forward in time, yielding  $T_{n + 1} = (s_t, a_t, r_{t+1}, s_{t+1}, \hat{a}_{t+1}, \hat{r}_{t+1}, \hat{s}_{t+2}, n=1).$ Then, we do a one-step update from $s_{t+1}$ to $\hat{s}_{t+2}$. 
$T_{n+1}$ is added back to $P$ and, if the TD-error is sufficiently
high, it will be popped off in the future to further extend the
trajectory from $\hat{s}_{t+2}$. In general, given a trajectory of $n$ steps we update at pair $(\hat{s}_{t+n}, \hat{a}_{t+n})$ with
\begin{align}
\delta = \hat{r}_{t+n+1} + \gamma \max_a Q(\hat{s}_{t+n+1},a) - Q(\hat{s}_{t+n}, \hat{a}_{t+n})
\end{align}
where the model is used to generate $\hat{r}_{t+n}$ and $\hat{s}_{t+n+1}$. 

This is one of the simplest strategies, and so variants of it are common \cite{Gu,Holland,Kalweit}. The original Dyna-Q algorithm \cite{Sutton1990} similarly uses a successor model and one-step updates, though it does not iterate the model (i.e., $\beta = 0$, which results in the priority of trajectories with $n \geq 1$ being $0$).

\textbf{Multi-Step Successor Dyna} uses a successor model with multi-step updates. A tuple ($s_t, a_t, \hat{r}_{t+1}, \hat{s}_{t+1}, \hat{a}_{t+2}, \hat{r}_{t+2}, \hat{s}_{t+2}, \ldots, \hat{s}_{t+n}$) is sampled from the queue $P$. Successor state $\hat{s}_{t+n+1}$ of $\hat{s}_{t+n}$ and reward $\hat{r}_{t+n}$ for some action $\hat{a}_{t+n}$ are generated. The agent updates at pair $(s_{t}, a_{t})$ with
\begin{align}
\delta = \sum_{i=0}^{n-1} \gamma^i \hat{r}_{t+i+1} + \gamma^n \max_a Q(\hat{s}_{t+n+1},a) - Q({s}_{t}, {a}_{t})
\end{align}

This multi-step strategy is less common than using one-step updates, but variants of the idea have been explored. \citeA{Yao} perform Dyna with a linear model and linear value function. They learn a multi-step model that directly predicts the expected reward and state multiple steps in the future, averaged over many timescales. 
\citeA{Buckman} also average over many multi-step updates, in this case weighted by a measure of uncertainty on the temporal difference error at each horizon. Outside of Dyna, multi-step updates are a common choice for Q-learning because value information can be quickly propagated.

\textbf{One-Step Predecessor Dyna} uses a predecessor model with one-step updates. The algorithm is similar to One-step Successor Dyna, but with updates using the reverse-dynamics trajectory. After $n$ iterations backwards with the model from an observed state $s_t$, we have a trajectory $\hat{s}_{t-n}, \hat{a}_{t-n}, \hat{r}_{t-n+1}, \hat{s}_{t-n+1}, \ldots, \hat{r}_{t}, s_{t}, a_t, r_{t+1}, s_{t+1}$. Then the predecessor model is queried, using $\hat{s}_{t-n}$ and $\hat{a}_{t-n-1}$, to get predecessor state $\hat{s}_{t-n-1}$ and reward $\hat{r}_{t-n}$, with complete transition $(\hat{s}_{t-n-1}, \hat{a}_{t-n-1}, \hat{r}_{t-n-1}, \hat{s}_{t-n})$. The agent updates at pair $(\hat{s}_{t-n-1}, \hat{a}_{t-n-1})$ with
\begin{align}
\delta = \hat{r}_{t-n} + \gamma \max_a Q(\hat{s}_{t-n},a) - Q(\hat{s}_{t-n-1}, \hat{a}_{t-n-1})
\end{align}
The canonical example of Dyna-style planning with predecessor state models is Prioritized Sweeping \cite{Peng,Moore}. The core idea of Prioritized Sweeping is, when a state is pulled off of the queue and its value is updated, its predecessors are added to the priority queue. \citeA{Sutton2008} used a small modification of this idea for linear models and value functions, by generating predecessor {\em features} for each state feature instead of  predecessor states. 
Recent work extends this approach to the function approximation setting \cite{Pan}, and one other recently illustrated the benefits of one-step predecessor models in the tabular setting \cite{Van}. One other work considered multi-step predecessor models, but instead used imitation rather than Dyna updates \cite{Goyal}.  

\textbf{Multi-Step Predecessor Dyna} uses a predecessor model with multi-step updates.
A tuple is sampled from the queue $P$, consisting of a reverse trajectory just like One-Step Predecessor Dyna. Then, a state $\hat{s}_{t-n-1}$ and reward $\hat{r}_{t-n-1}$ for action $\hat{a}_{t-n-1}$ are generated. The agent updates at pair $(\hat{s}_{t-n-1}, \hat{a}_{t-n-1})$ with
\begin{align}
\delta = \sum_{i=0}^{n} \gamma^i \hat{r}_{t-n+i} + \gamma^n \max_a Q({s}_{t},a) - Q(\hat{s}_{t-n-1}, \hat{a}_{t-n-1})
\end{align}
Somewhat surprisingly, this fourth variant has not been previously considered. In this work, we explore this understudied region of the Dyna design space. The key feature of Multi-step Predecessor Dyna is that the agent only updates towards the values of real states, $\max_a
Q(s_t, a)$. Thus, we hypothesize that Multi-step Predecessor state Dyna will be more robust to model error than the other three variants discussed above, because it uses real state values in the target rather than the values of simulated states. 

It is worth pointing out that there is also one other important benefit of Multi-Step Predecessor models when using target networks. A target network is a fixed estimate of the action-values, for use in the bootstrap target, that is periodically updated to the current action-values. Since target networks are updated with lower frequency compared to the action-values, they do not immediately reflect the action-value modifications. Therefore, they might slow down how reward information is propagated backwards under one-step updates, whereas multi-step updates avoid this issue. 

To better understand why, consider the case where One-step Predecessor Dyna updates backwards from a goal state, with observed reward $r_{t+1}$ from state $s_t$ after taking action $a_t$. 
The action-value $Q(s_t,a_t)$ is updated using $\delta = r_{t+1} + 0 - Q(s_t, a_t)$. However note that the target network $\tilde{Q}$ is not updated. Now when we iterate one step backwards, we generate predecessors $\hat{s}_{t-1}, a_{t-1}$ with predicted reward $\hat{r}_{t}$ and update with
$\delta = \hat{r}_{t-1} + \gamma \max_{a'} \Tilde{Q}(s_t, a') - Q(\hat{s}_{t-1}, a_{t-1})$. The information from the goal state is not propagated to $Q(\hat{s}_{t-1}, a_{t-1})$ because we bootstrap off the target network rather than the updated values. However, for Multi-step Predecessor Dyna, we directly update with the generated rewards and bootstrap off only the terminal state, namely we use  $\delta = [\hat{r}_{t-n+1} + \gamma \hat{r}_{t-n+2} + \ldots + \gamma^{n-2} \hat{r}_{t} + \gamma^{n-1} r_{t+1} + 0] - Q(\hat{s}_{t-n}, {a}_{t-n}).$

\begin{table*}[t]
	\caption{Previous work corresponding to the four Dyna variants. Multi-step Predecessor Dyna has not been previously proposed.}
	\label{Table:Section4/Algo_Quadrants}
	\vskip 0.1in
	\begin{center}
		\begin{small}
			\begin{tabular}{>{\columncolor[HTML]{EFEFEF}} p{1.61cm}| p{6.27cm}p{6.27cm}} 
				\toprule
				& \cellcolor[HTML]{EFEFEF}One-Step Updates & \cellcolor[HTML]{EFEFEF}Multi-Step Updates \\ \midrule
				\\
				Successor & \textbf{One-Step Successor}  \cite{Sutton1990}, \cite{Holland}, \cite{Gu}, \cite{Kalweit} & \textbf{Multi-Step Successor} \cite{Peng}, \cite{Moore}, \cite{Sutton2008}, \cite{Goyal}, \cite{Pan}         
				\\
				\\
				Predecessor & \textbf{One-Step Predecessor}
                                \cite{Yao}, \cite{Buckman} & \textbf{Multi-Step Predecessor} Introduced in this work.
			\end{tabular}
		\end{small}
	\end{center}
\end{table*}

\section{The Hallucinated Value Hypothesis}\label{Section:Hypothesis}
The planning procedure in Dyna-style algorithms can be error-prone due to bootstrapping from wrong value estimates. Even in model-free RL, bootstrapping can incur error due to having an inaccurate value function. This issue is exacerbated in planning, when the model might produce erroneous next states or rewards. In this section, we discuss why this is the case, by introducing the hallucinated value hypothesis, and why Multi-step Predecessor Dyna helps mitigate the issue.

Consider the case where an imperfect model produces a next state that is unreachable, or non-existent. This model error can harm performance during planning, because the agent bootstraps off the value in this unreachable state. The value of this states is never updated directly from real experience because that state is not reachable by the agent. Rather, the values of such states are essentially arbitrary, depending on value function initialisation and how the function approximator generalises. As a result, when updates are performed using simulated experience, the value of real states may be contaminated by these arbitrary values. In turn, this may mislead the control policy. 

This issue can arise even if states are reachable, but are rarely visited. The values for such states can be similarly arbitrary, because they have not been updated sufficiently, and so we call them hallucinated. If the model produces a transition to such a rare state, and the agent bootstraps off of these hallucinated values, then again it can skew the action-values. Models that are iterated multiple steps are more likely to move further away from states that are frequently visited, or produce non-existent states.

We propose the following hypothesis.
\vspace{-0.1in}
\begin{quote}\it
\textbf{Hallucinated Value Hypothesis:}
	Planning updates that result in the values of real states being updated towards the values of simulated states impedes learning of the control policy.
\end{quote}

\subsection{BorderWorld: A Motivating Example for the HVH}
\begin{figure}
	\centering
	\subfloat[]{\includegraphics[width=0.35\columnwidth]{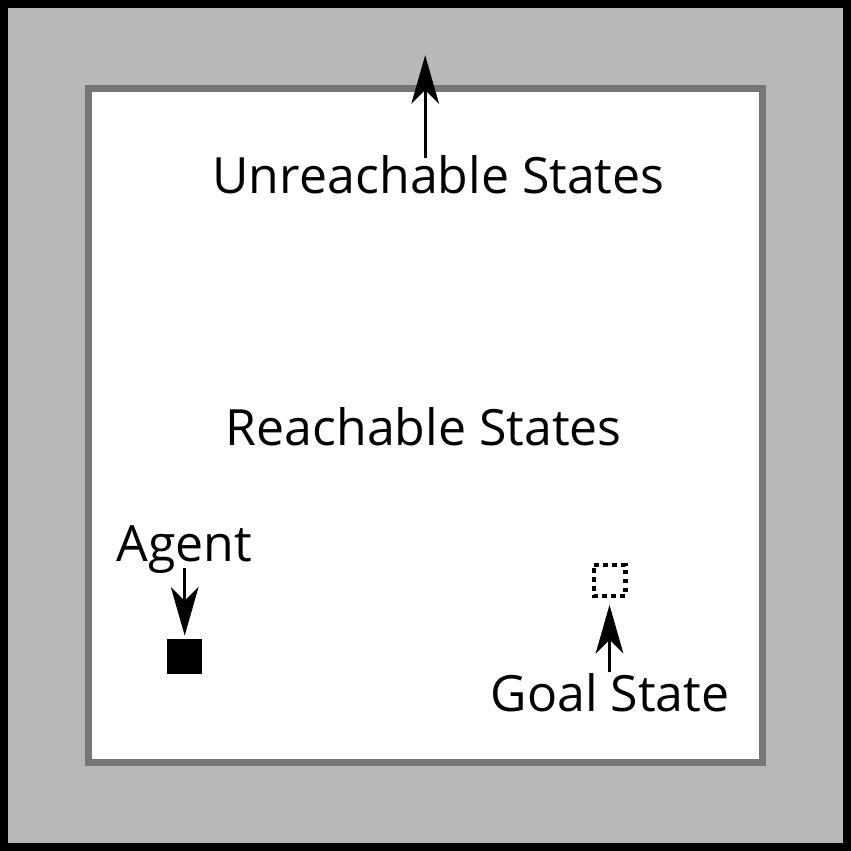}}
	\hspace{1em}
	\subfloat[]{\label{Figure:Section2/BorderedGridworld}\includegraphics[scale=0.8]{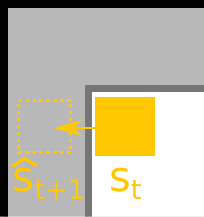}}
	\caption{(a) The Borderworld environment. (b) An example of an erroneous simulated transition in BorderWorld.}
	\label{Figure:Section2}
\end{figure}

{\label{Section:BW_motivating_example}}
In this section we present an example explaining and motivating the HVH. Figure \ref{Figure:Section2} (a) shows Borderworld, a 2D navigation environment. The state is represented by the tuple $(x,y)$ indicating the agent's position, and the actions North, East, South, West result in the agent moving in the respective cardinal direction. The reward is $1$ on transitioning into the goal state and $0$ elsewhere. The key feature of Borderworld is a border of unreachable states. These states form a set $\mathcal{S}_{U}$, while the reachable states form a set $\mathcal{S}_{R}$. In the underlying MDP there are no transitions from $\mathcal{S}_{U}$ to $\mathcal{S}_{R}$. 

However, imperfect models may predict such transitions, particularly due to the generalization that naturally occurs in $(x,y)$ space.
Suppose a neural network is trained to predict transition dynamics on Borderworld. In most states taking an action results in a deterministic change to the agent's position. For example, taking the action West from $(3, 1)$ results in the agent's position changing to $(2, 1)$. The dynamics of West are similar across most of the environment, and so a neural network will likely generalise this behaviour even to states beside the border: it may generate simulated state $(\hat{x}, \hat{y})$ where $\hat{x}$ is in the border, as we visualize in Figure \ref{Figure:Section2} (b). 

Now the question is if such a small error really impacts the performance of a Dyna-Q agent.
Consider the transition in Figure \ref{Figure:Section2} (b). The value
of $Q(s_t, \text{West})$ is updated towards the target $\hat{r}_{t} +
\gamma \max_a Q(\hat{s}_{t+1}, a)$. However, this target can be
misleading because the value $Q(\hat{s}_{t+1}, a)$ is not updated with
real experience and is essentially arbitrary. Suppose the action-value function is optimistically initialised so that unvisited states have high value. The value of $\max_a Q(\hat{s}_{t+1}, a)$ will be large, which will in turn raise the value of $Q(s_t, \text{West})$. This may cause the agent to prefer moving to the wall, rather than moving toward the goal. Furthermore, the erroneous high value of the simulated state cannot be corrected through real experience as the agent cannot reach $\hat{s}_{t+1}$ to change its value.

This example highlights that Dyna algorithms which update real state values towards simulated state values may be brittle in the face of model error. In a simple example such as Borderworld, the agent may eventually correct its model error; it is, after all, driven to experience states where its model is incorrect. However, in more complex environments it is unreasonable to expect the model to correct all of its errors. In Section \ref{Subsection:Experiments/ReinforcemenLearningBenchmarks} we observe negative effects from hallucinated values in several standard benchmark domains. We also see that these errors are persistent and do not resolve on their own.

\subsection{HVH in the Four Variants of Dyna} \label{hvh4variants}
Now that we have developed a design space of Dyna algorithms in Section \ref{Section:ADynaDesignSpaceOfDynaAlgorithms}, let's discuss their implications with respect to the HVH.

\textbf{One-Step Successor Dyna and the HVH:}
We can reason about how this approach performs in Borderworld. The problematic transition discussed in Section \ref{Section:BW_motivating_example} will be generated --- the value of moving toward the border will be erroneously increased as a result of the high value of an unreachable border state. When iterating the model, however, this error may be corrected over time. 
The model may simulate transitions from border states back to reachable states.
Since the model is used to simulate multiple steps, it may not only simulate transitions from reachable states to border states, but also from border states back to reachable states. 
As such, the values of these unreachable states are updated during planning. Thus, in Borderworld, it may be possible that hallucinated values eventually cease to be an issue, but this may take a great deal of time, likely harming rather than helping sample complexity.

\textbf{Multi-Step Successor Dyna and the HVH:}
Hallucinated values might be even more of a problem in Borderworld under multi-step updates with successor models. The agent can still generate trajectories into the border and update values for real states towards the hallucinated values of the border states. Unlike one-step updates, however, the values of these border states will not be corrected as the multi-step update only updates \emph{real} state values. This occurs because the multi-step update progressively increases the horizon, $n = \{1, 2, 3, ...\}$. For the first planning step $s_{t}$ gets updated towards $s_{t+1}$, for the second planning step $s_{t}$ gets updated towards $\hat{s}_{t+2}$, and so on. At no point is the value for a simulated state $\hat{s}_{t+n}$ updated. Note that if we used a fixed $n$ for the $n$-step target, rather than a growing $n$, then we would use a sliding window on the generated trajectory and so simulated states would be updated.

\textbf{One-Step Predecessor Dyna and the HVH:}
In this algorithm the value of $\beta$ plays a key role with respect to whether it suffers from the problem posed in the HVH. If $\beta > 0$, then this approach can still suffer from hallucinated values, as it
updates values of real states using simulated states. For example,
if the agent is in state $s_t = s$ beside the
border, it can generate $\hat{s}_{t-1}$ outside the border and then
$\hat{s}_{t-2} = s$ back inside the border. Consequently, when it
updates $\hat{s}_{t-2}$ using $\hat{s}_{t-1}$, it will update the value of a real state using a simulated state. However, if we
prevent it from extending the backwards trajectory further from
$\hat{s}_{t-1}$, after the first iteration, there is no chance of
updating the value of a real state towards the value of a simulated
state. This precisely occurs when $\beta=0$ (and the
priority threshold $\rho > 0$), as the priority for the iterated transition will be zero and it will not be added to the queue and so will not be iterated further. All planning updates under such a setting, therefore, only update values of simulated states to
values of real states. We call this version of the algorithm, where $\beta = 0$, Uniterated One-step Predecessor Dyna and hypothesize that it should be less prone to the hallucinated values.\footnote{\citeA[Section 2.3]{Van} also hypothesized that updating simulated states might be less error-prone than updating real states with the values of simulated states.}

\textbf{Multi-Step Predecessor Dyna and the HVH:}
We claim that this algorithm is more robust to model error since it only updates state values toward a real state. Similar to the reasoning mentioned for One-step Predecessor Dyna, consider a state beside the wall in the BorderWorld. The agent starts in state $s_t = s$ beside the
border and then generates $\hat{s}_{t-1}$ outside the border. It updates $\hat{s}_{t-1}$ according to the value of a real state $s_t = s$. This simulated transition would be added to the trajectories to be iterated further. By rolling out such a trajectory from $\hat{s}_{t-1}$, we go to $\hat{s}_{t-2}$, which also could either be inside the borders or outside. The key point is that in either situation, Multi-step Predecessor Dyna updates the value of $\hat{s}_{t-2}$ from a real state $s_t = s$.

Note that this property is a consequence of using multi-step updates that grow as the trajectory is iterated. The agent updates first with a one-step update, then extends the trajectory, so that next time it updates with a two-step update. This process continues, updating with a three-step update the next time, and so on. This contrasts using a fixed $n$-step update. If the trajectory reached a length of $n+1$, then the agent would only use a subset of the trajectory for the update. In particular, it would no longer use the real state at the end of the trajectory, and would possibly instead bootstrap off of a simulated state, namely use $Q(\hat{s}, \dot)$ in the bootstrap target from simulated state $\hat{s}$.
This is one of the reasons we chose for a variable length multi-step update, in addition to simplicity.

\subsection{The Impact of Generalization when Using Simulated States}
In this section, we reason further about the impact on value estimates when updating with simulated states. We discuss why updating simulated states towards the values of real states should be less problematic, even though generalization could still impact real states. 

\begin{figure}[t]
	\centering
	\includegraphics[scale=0.45]{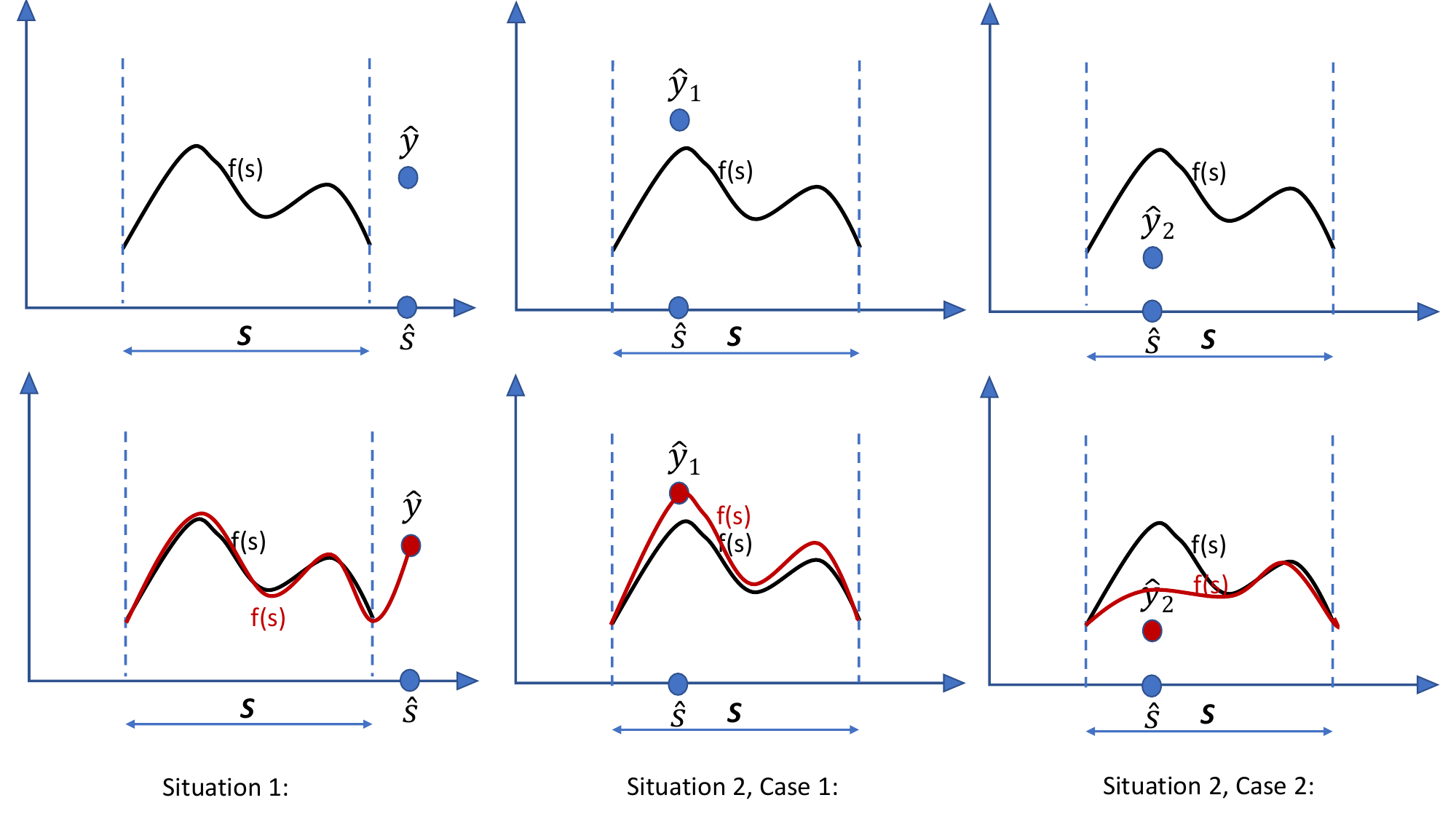}
	\caption{\textbf{Situation 1} corresponds to updating on a simulated state that is far from the experienced states in the environment. In this situation, a simulated state $\hat{s}$ is less likely to skew $f(s)$ when it is updated with ($\hat{s}$, $\hat{y}$). \textbf{Situation 2} corresponds to updating a (simulated) state within the experienced states $\mathcal{S}$. In this situation, the given target might skew $f(s)$ (Case 2) or not (Case 1). 
	}
	\label{Figure:HVH_theory}
\end{figure}

Let us consider a setting where we learn a function $f(s)$, using sampled inputs (states) $s$ and targets $y$. For us, this will correspond to the states from which we update in Dyna, where $y$ is the TD target in the planning step that is different for each Dyna variant. When learning $f$, we can imagine obtaining inputs $\hat{s}$ outside the range of the set of states we have seen so far---representing only simulated states---labeled Situation 1 in Figure \ref{Figure:HVH_theory}. The other setting, which we label Situation 2 in Figure \ref{Figure:HVH_theory}, is when the training input $\hat{s}$ is within the range of observed states, sometimes called in-distribution. Situation 2 could arise with either real states or simulated states. For this situation, there are two cases: either the corresponding target is consistent with the current function $f$ and only minorly changes the function (Case 1) or the target is very different and causes a large change in the function (Case 2). 

Situation 1 corresponds to the case where we update a simulated state. The only Dyna variant that does not update simulated states is Multi-step Successor Dyna, which always updates real states towards multi-step bootstrap targets computed from simulated experience. Situation 1 can affect the function $f$ as well, but with sufficient capacity, it can simply locally fit the observed $(\hat{s}, \hat{y})$ without impacting the function for real states that have been observed. 

The most problematic setting is Situation 2, Case 2. This occurs when we update states within the range of states we have seen, but have generated a target that heavily skews our existing function. This could occur when the target uses hallucinated values and we are updating a real state. The agent could even have relatively accurate action-values from real experience, that could then be skewed when updating with this simulated transition. Situation 2, Case 1 is a setting where the model produces relatively consistent transitions with the real-world, and so does not cause issues. 

Multi-step Predecessor Dyna is more likely to be in Situation 2, Case 1. The agent always bootstraps off the values of real states, and so we expect that the target $\hat{y}$ should be relatively consistent with the function. Multi-step Successor Dyna, on the other hand, is likely to suffer from Situation 2, Case 2, when iterating the model forward and bootstrapping off of simulated states. One-Step Predecessor Dyna and One-step Successor Dyna should be slightly less prone to this bad outcome. Though they can bootstrap off of the value of simulated states, they can at least make the function more consistent by updating the values for those simulated states. However, they do still bootstrap off of the values for simulated states, which can cause bootstrap targets that are inconsistent with bootstrap targets on real-experience. 

One other benefit of applying multi-step updates rather than one-step updates is that they can balance between reward and action-value function approximation errors. One-step methods rely heavily on the action-value estimates, since they bootstrap immediately. As we increase the number of steps, we rely more on the learned reward function that generates the sequence of rewards. Generally speaking, neither the smallest nor the largest number of update steps (represented by $\beta$ in our case) yield the best learning performance \cite[Chapter 7]{sutton2018reinforcement}. Section \ref{Section:beta} explains the role of $\beta$ in the number of update iterations in detail.
Further, Multi-step Predecessor Dyna may enjoy another benefit, when bootstrapping off of values near the goal in episodic tasks. The action-values near the goal are potentially more accurate, as the horizon is shorter and so estimation is simpler. The multi-step approach with predecessor models more often bootstraps off of these states, than the other three variants.

\section{Experiments}\label{Section:Experiments}

In this section, we report several experiments to investigate the HVH, for the four Dyna variants.
We first conducted controlled experiments on Borderworld, which show clear evidence supporting the HVH: the Dyna algorithms that update values of real states to values of simulated states all fail while the algorithms that do not perform such planning updates succeed. 
We then conducted similar experiments on three RL benchmarks including Catcher, PuddleWorld, and CartPole, where their results also support the HVH. We finally concluded by demonstrating that the HVH persists even when allowing the model to update online. 

Throughout these experiments, we use the multi-step update that uses on-policy sub-trajectories. We discuss this choice and compare it to the more biased approach of using the immediate off-policy update, without further trajectory expansion, in Appendix \ref{appendix A}.

\subsection{A Controlled Experiment on Borderworld}\label{Subsection:Experiments/Borderworld}

Planning updates are problematic if the environment model produces unreachable states as updating toward these states propagates arbitrary value. To mimic these conditions in Borderworld we made two design choices: first, we introduce a flaw in an otherwise perfect environment model of Borderworld. Specifically, the model generates transitions from real states to (unreachable) border states, as in Figure \ref{Figure:Section2/BorderedGridworld} (and from border states to border states). Second, we optimistically initialised the value function to $1$ so that border states would have misleading values; planning updates for actions leading into these states would increase.

We used the algorithms as specified in Section \ref{Section:ADynaDesignSpaceOfDynaAlgorithms} with a tabular value function. We compared the four Dyna variants, as well as the Uniterated One-step Predecessor. We include the Uniterated variant, which corresponds to using $\beta = 0$ and so does not iterate backwards. It serves as a baseline to remove confounding factors related to whether differences were due to hallucinated values or using multi-step updates. One-step Successor, Multi-step Successor, and  One-step Predecessor update toward simulated states and could suffer under the HVH. Multi-step Predecessor and Uniterated One-step Predecessor both only update towards real states.

The agents are run for 250000 steps in Borderworld of size $42 \times 42$, with learning curves showing accumulated reward. All curves are averages over $30$ runs and for the best hyperparameters for each algorithm. The value of $\alpha$ has been selected by sweeping over a set of \{0.1,  0.25, 0.5, 0.75, 0.05, 0.125\} and $\beta$ by sweeping over \{0.0, 0.15, 0.33, 0.50, 0.66, 0.75, 0.90, 1.0\}. 
All agents use $N = 1$ planning updates per step, where each planning update iterates over all actions. Note that all the one-step methods get the exact same number of updates: $N |\mathcal{A}|+1$ updates per-step, one for the online update and $N |\mathcal{A}|$ planning updates. The multi-step algorithms get almost the same number of updates, but because the screening approaches periodically drop off-policy trajectories, at times they do not update because the queue is empty. We recorded the number of updates and found them to be sufficiently similar (only differences of up to a few hundred updates).

\begin{figure}
	\centering
	\includegraphics[width=0.65\linewidth]{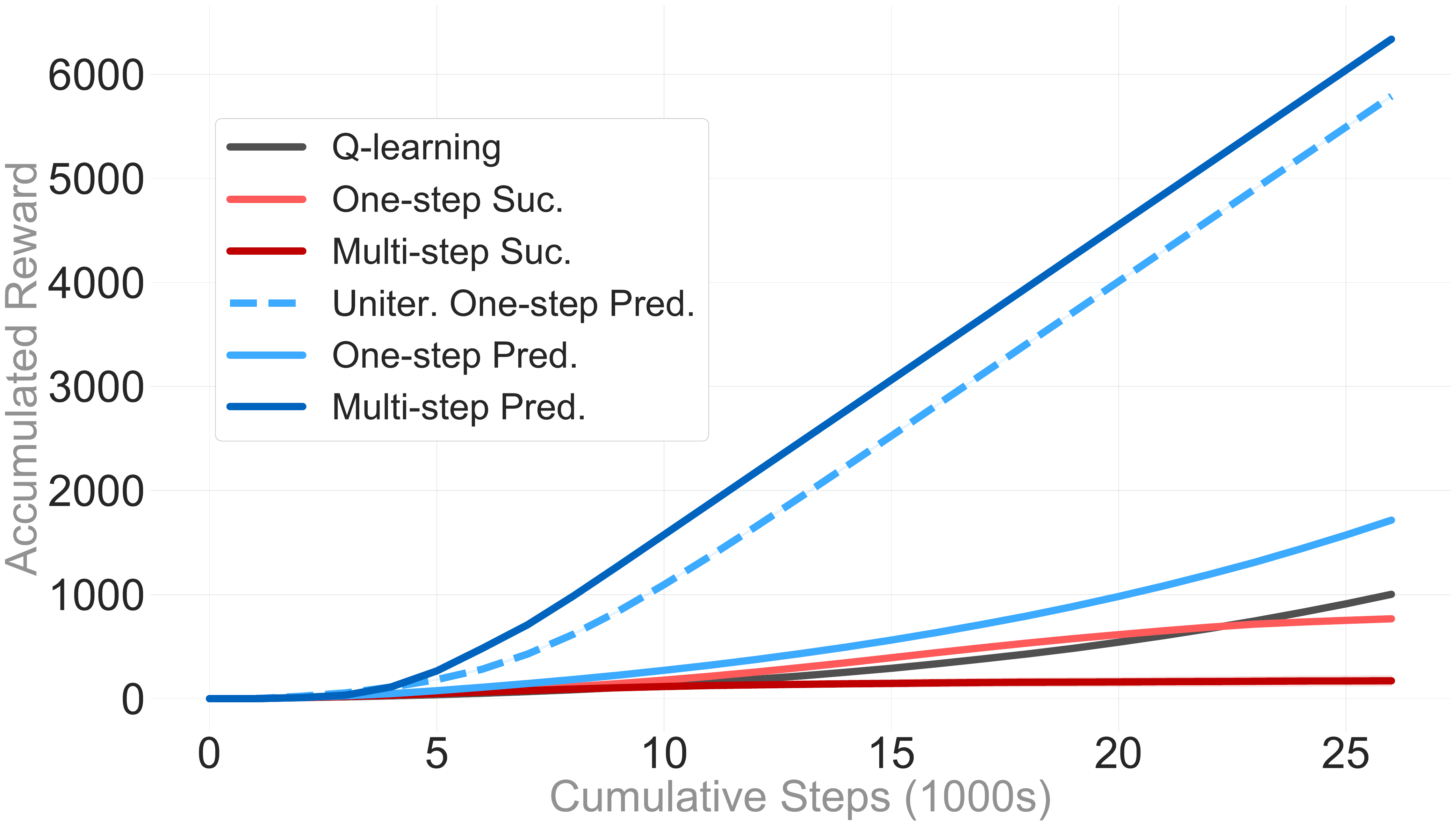}
	\caption{Learning curves on Borderworld over all of the algorithms with the same computational complexity and approximately the same number of updates. Multi-step Dyna variants allow biased one-step updates similar to One-step variants, but would not put that unbiased transition back into the queue for further rollouts. The screening approach in Multi-step variants is also only possible on the on-policy sub-chunks of a trajectory to avoid biased TD errors as much as possible. Error bars are not visible as they are smaller than line thicknesses.}
	\label{Figure:Section5/BW_Learning_Curves}    
\end{figure}

\begin{figure}
	\centering
	\includegraphics[width=0.75\linewidth]{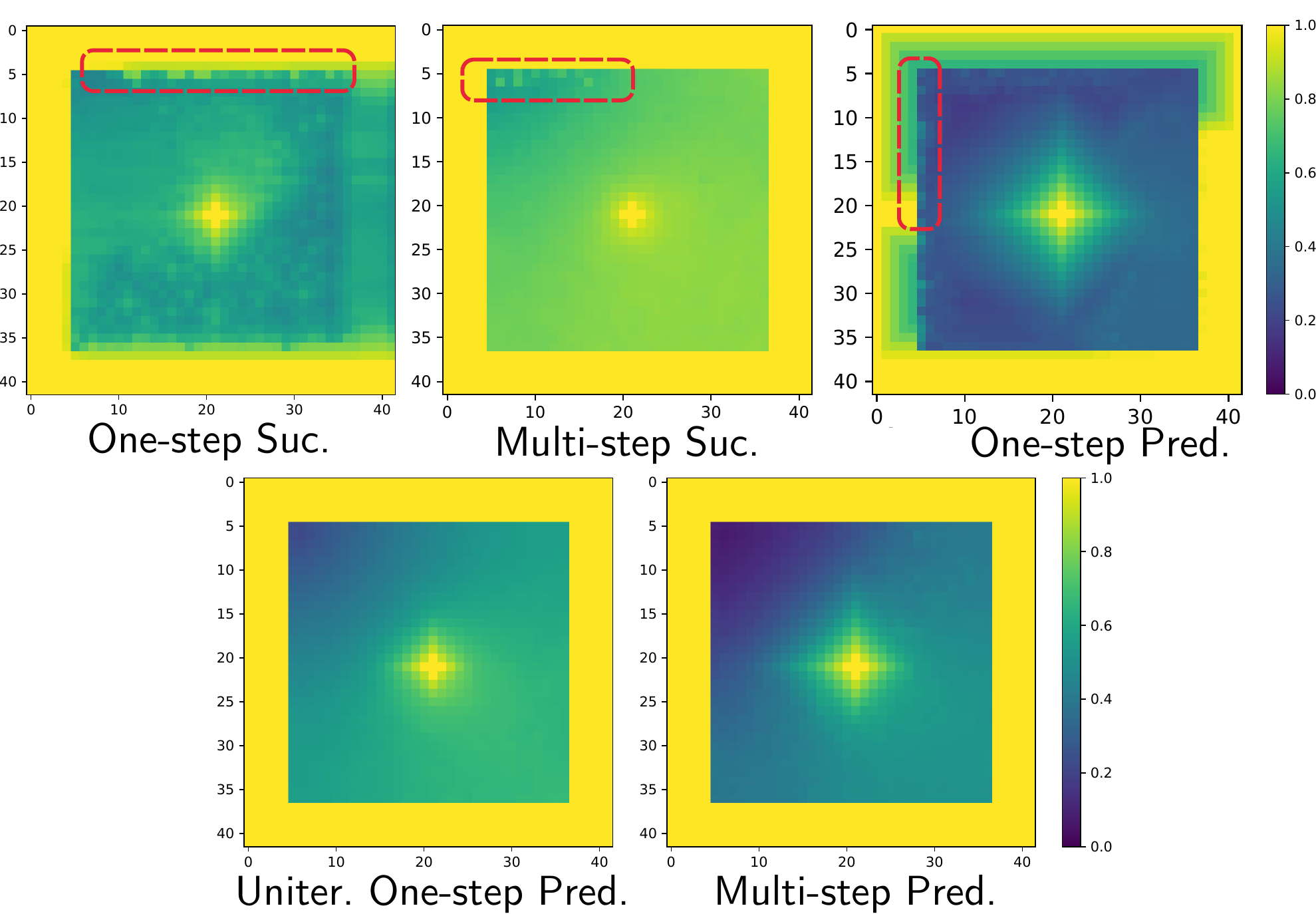}
	\caption{Plot of $\max_a Q(s, a) \:\forall s \in \mathcal{S}$ after $100,000$ steps. The red rectangles show where values of real states have been contaminated by values of simulated states.}
	\label{Figure:Section5/BW_Heatmaps}    
\end{figure}

Figure \ref{Figure:Section5/BW_Learning_Curves} shows that the methods that update towards simulated states perform notably worse than those that do not. The learning
curves for One-step Successor,  Multi-step Successor and Iterated One-step Predecessor 
increase slightly early on but then flatten out. Heatmaps of $\max_a Q(s, a)
\:\forall s \in \mathcal{S}$ after $100,000$ steps for the three
failing algorithms are shown in Figure
\ref{Figure:Section5/BW_Heatmaps}. The plots for these three
approaches (top row) all show high values for reachable states near the border. In Borderworld, the only transitions with real rewards are those that lead to the goal state in the centre. Therefore, the high values near the border must be the result of planning updates propagating values from border states to real states near the border. When the agent reaches one of these states close to the border instead of taking actions that move it closer to the goal, it takes actions toward the border, chasing hallucinated value. In contrast, Uniterated One-step Predecessor and Multi-step Predecessor do not get stuck in this suboptimal behavior, and accumulate significantly more reward. 
Indeed, in Figure \ref{Figure:Section5/BW_Heatmaps}, we do not see contamination of
values of real states.

One interesting phenomenon shown in Figure
\ref{Figure:Section5/BW_Heatmaps} is that for One-step Successor and One-step Predecessor, the values of border states are lower
than their initialisation values. For these algorithms, after sufficient planning, hallucinated values may be updated to be more similar to those of real states---eventually, they might no longer mislead the agent. However, this may take a long time and the agent will be catastrophically misled in the meantime. 

In the case of Multi-step Predecessor and Multi-step Successor, we expect to see the same effect on the border's action values since the planning procedure should let the border states get occasional updates towards the real states' action values; however, we barely see any changes. This is due to the screening method applied during planning, where almost all trajectories that would update in the border are thrown away.

To describe the screening procedure for Multi-step Predecessor methods, imagine we start with the last transition $(s_{t-1}, a_{t-1}, s_{t})$, with both $s_t$ and $s_{t-1}$ real states right beside the border. Rolling out further backwards, we get $(s_{t-2}, a_{t-2}, s_{t-1}, a_{t-1}, s_{t})$ with $s_{t-2}$ potentially being outside the border with a high hallucinated initial value. Likely, action $a_{t-2}$ is non-greedy because it goes from the high-valued border state $s_{t-2}$ into the lower-valued real state $s_{t-1}$, while the other three possible actions go to other high-valued surrounding border states. 
This means that $Q(s_{t-2,}, a_{t-2,})$ gets updated but is not put back in the priority queue for further roll-out. 
Therefore, border states do get updated, but much less frequently to the point that is not visible to us in Figure \ref{Figure:Section5/BW_Heatmaps}.

Similarly, for Multi-step Successor methods, imagine we start with the last transition $(s_{t}, a_{t}, s_{t+1})$, with a real state $s_t$ beside the border and a border state $s_{t+1}$. The screening method checks whether the action $a_t$ is greedy or not. Likely, $a_t$ is greedy due to the highly-valued border states. The value of near-border states $s_t$ gets contaminated from highly-valued border state $s_{t+1}$ and the trajectory is put back in the priority queue. Rolling out further forwards, we might get a looped-back trajectory from border to the near-border states such as $(s_{t}, a_{t+1}, s_{t+1}, \cdots, s_{t+\tau-1}, a_{t+\tau-1}, s_{t+\tau})$ with $s_{t+\tau-1}$ a highly-valued hallucinated state in the border and $s_{t+\tau}$ with a near-border real state. Likely, action $a_{t+\tau-1}$ is non-greedy because it goes from the high-valued border state $s_{t+\tau-1}$ into the lower-valued real state $s_{t+\tau}$, while the other three possible actions go to other high-valued surrounding border states. 
This means that $Q(s_{t+\tau-1}, a_{t+\tau-1})$ gets updated but is not put back in the priority queue for further roll-out. 
Therefore, border states barely get updated due to the screening method and the fact that the looped-back trajectories are rare.

\subsection{Experiments on Reinforcement Learning Benchmarks}\label{Subsection:Experiments/ReinforcemenLearningBenchmarks}
In this section, we investigate whether hallucinated values play a role in a typical benchmark setting, where we have not artificially introduced conditions that should cause it to occur. 
First, we did not artificially introduce errors to the environment
model. Rather, we learned an environment model from data, which naturally has some error.
Second, we did not place any limitations on the initialisation of the
agent's weights. 

Our experiments were conducted on three benchmarks: \emph{Cartpole} \cite{Brockman}, \emph{Puddleworld}
\cite{Degris}, and \emph{Catcher} \cite{Tasfi}. For each algorithm, we
swept over $\alpha$ to pick $10$ randomly selected from a proper range 
in [0, 0.5]
and the performance of each hyper-parameter setting was averaged over $30$ random seeds. 
The selected learning rate set for Cartpole and Puddleworld is \{0.045, 0.065, 0.085, 0.095, 0.12, 0.15, 0.07, 0.205, 0.075, 0.08\}. For Catcher, we also swept over 10 smaller numbers in the range of [0, 0.05], which are \{0.0045, 0.0065, 0.0085, 0.0095, 0.0012, 0.0015, 0.007, 0.0005, 0.0025, 0.008\}. 
We reported performance for the best hyperparameter choice for each algorithm. 

\textbf{Learning Environment Models.} 
We ran two sets of experiments: one with an offline learned model and another where the model is allowed to adjust online. For the online setting, we loaded the pre-trained environment model and continued training the models on data gathered online while the agent was acting.
To learn the offline model, following the method of \cite{oh2015action}, we collected $100,000$ training samples by executing a pre-trained agent on the environment with $\epsilon = 0.5$. The offline model itself is a neural network (NN), with input state $s_t$ as well as a $1$-hot encoding of action $a$. The network outputs a vector $\hat{y}$ consisting of the next or previous state $\hat{s}_{t\pm1}$ and reward $\hat{r}_{r\pm1}$  ($\pm$ indicates we may be modelling forward or backward dynamics). The network was trained to minimise mean-squared error (MSE) between the prediction $\hat{y}$ and ground truth $y$, $\frac{1}{k}\sum_{i=0}^k (\hat{y_i} - y_i)^2$.

\textbf{Learning Value Functions.} We used a linear function approximator to
learn the action-values. To generate state features, we used the activation
of the final hidden layer of a pre-trained DQN \cite{Mnih} agent. We
trained a network with $200$ hidden units to convergence using the DQN
algorithm and froze its weights. In each step we input state $s_t$ to
the network and extracted the hidden layer activation to form a vector
of state features $\phi(s_t)$. The value function was linear in
$\phi(s_t)$. We initialised weights of the linear learner using
samples from $\mathcal{N}(0, 1)$. This ensures that states have a
variety of initial values, which may be optimistic or pessimistic.

\begin{figure}[t]
	\centering
	\subfloat[\emph{Cartpole}]{\includegraphics[width=0.5\columnwidth]{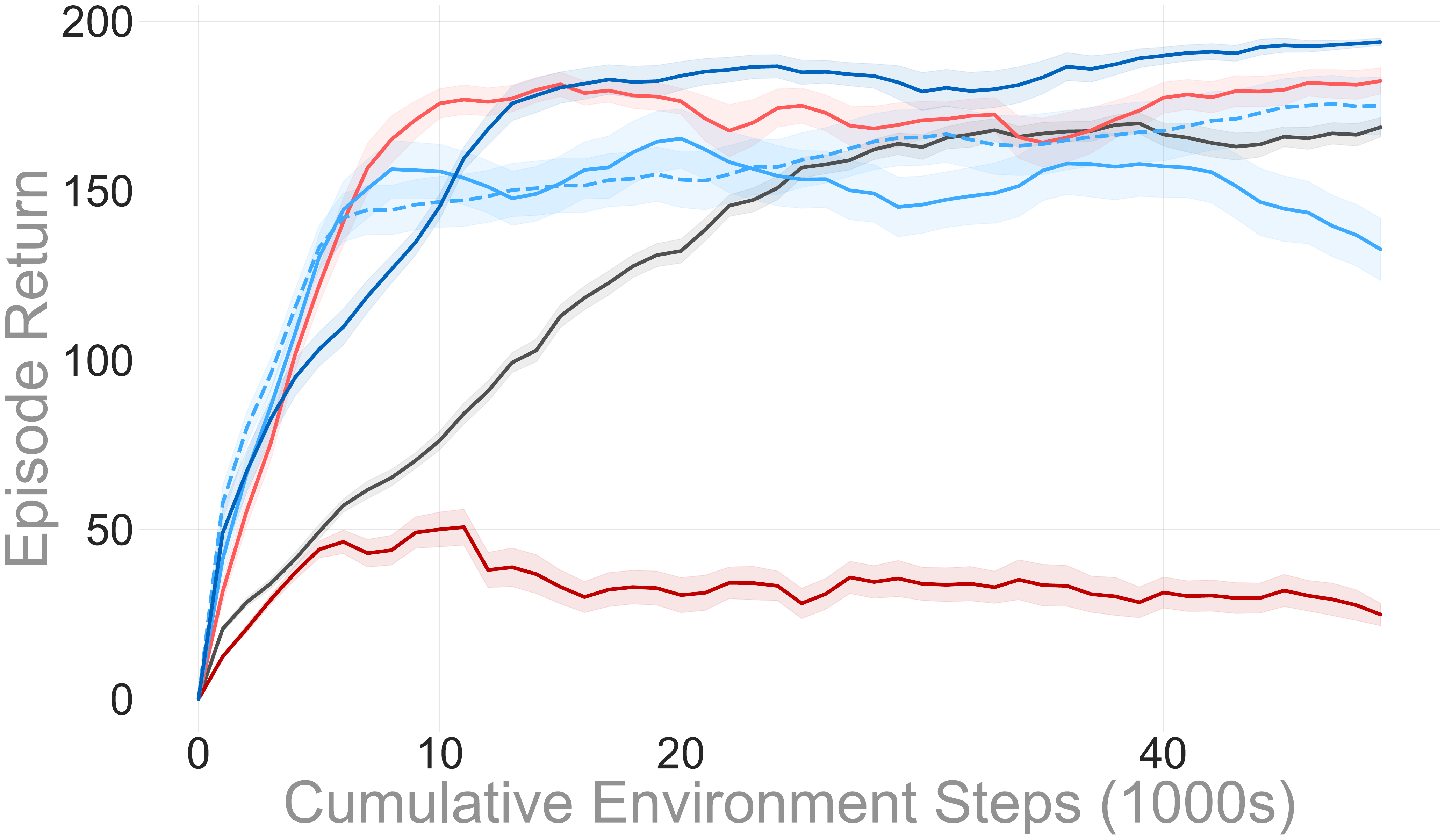}}
	\subfloat[\emph{Catcher}]{\includegraphics[width=0.5\columnwidth]{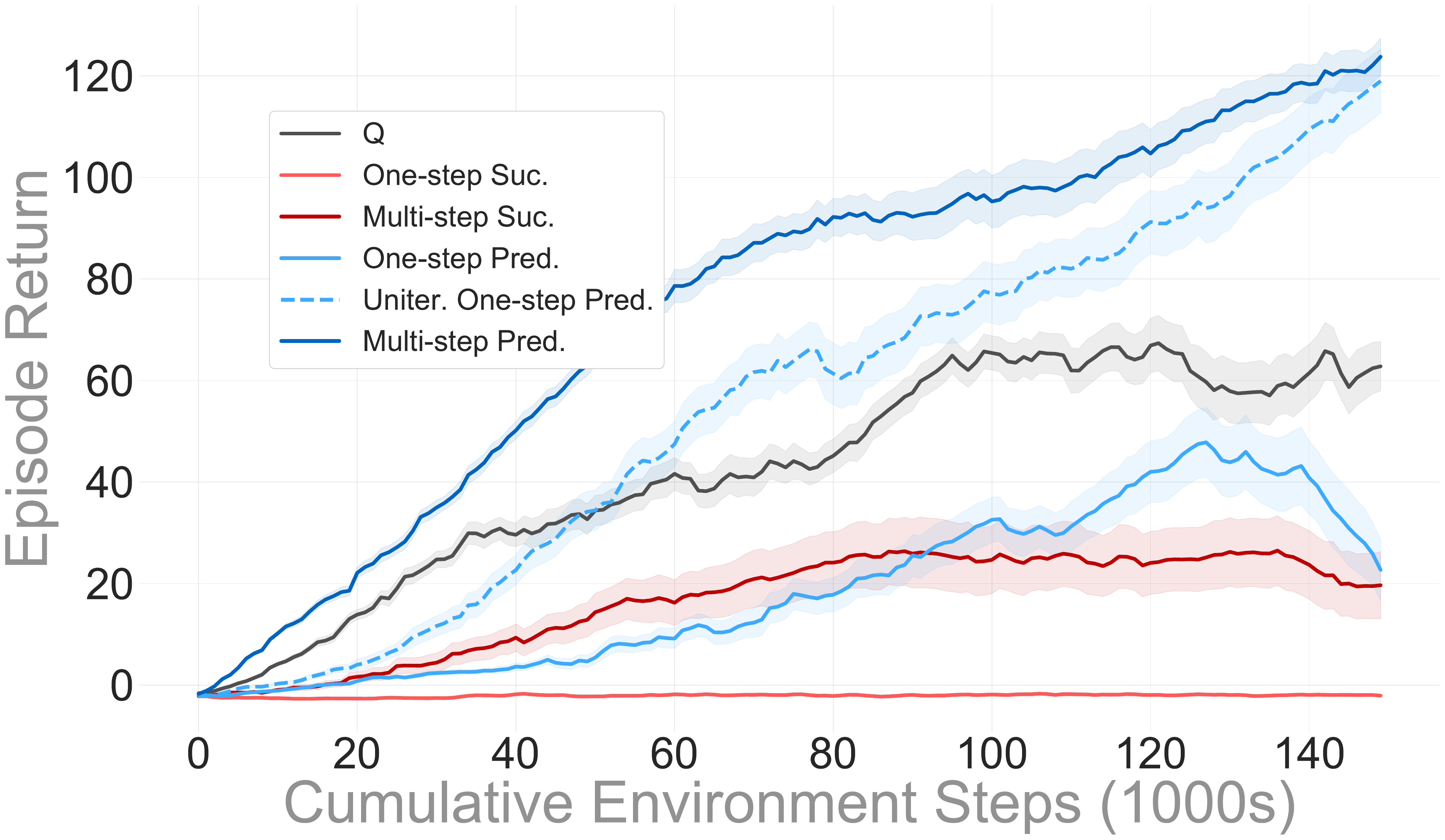}}\\
	\subfloat[\emph{Puddleworld}]{\includegraphics[width=0.5\columnwidth]{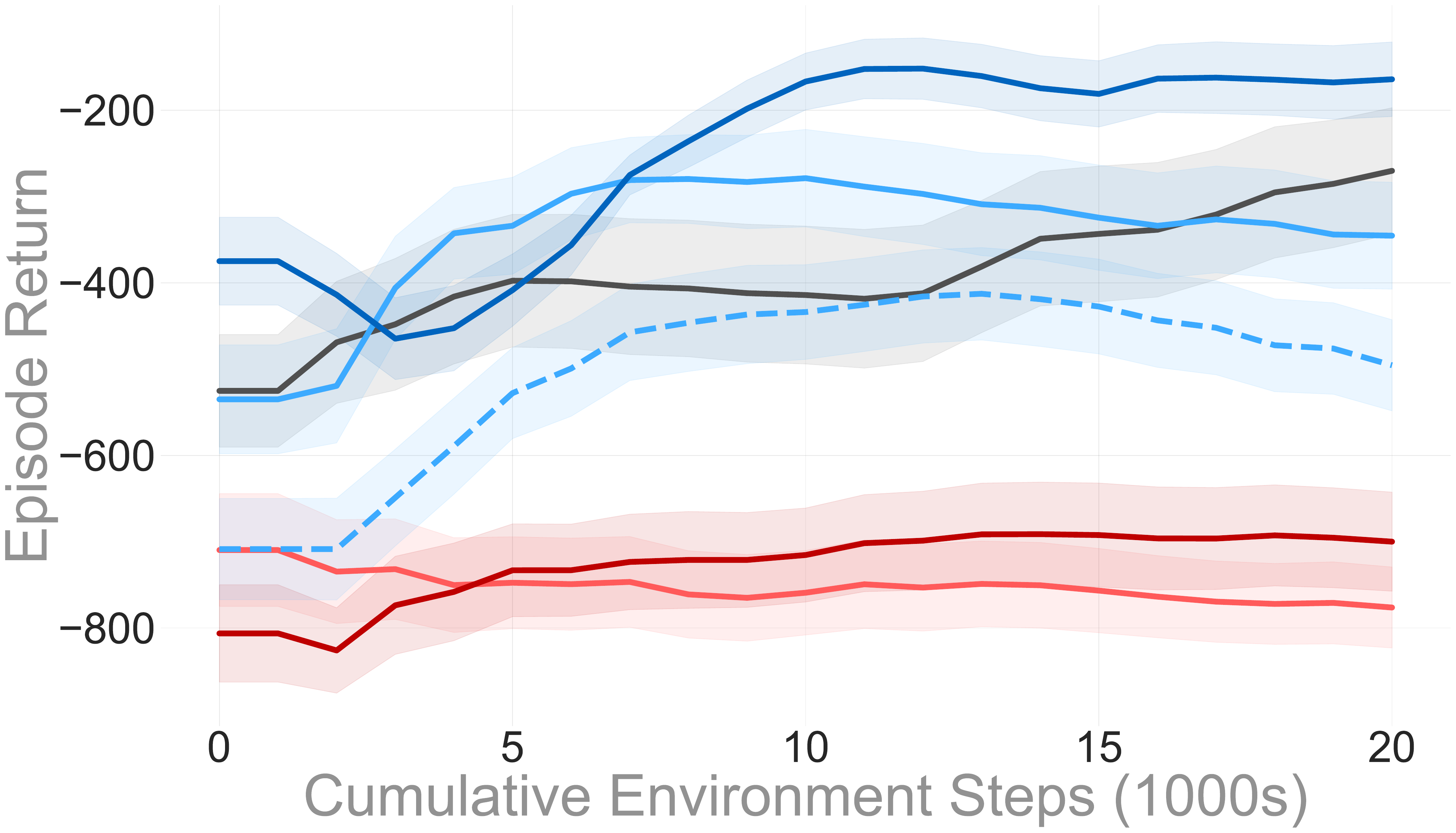}}
	
	\caption{Learning curves for the Dyna variants with a fixed environment model learned \textbf{offline}. All curves are averaged over 30 runs, with shaded regions corresponding to standard errors. The blue lines are predecessor models and the red lines are successor models. We can see generally that Dyna with predecessor models is more effective, and that the variants that Multi-step Predecessor performs significantly better in all environments.}
	\label{Figure:Section6/LearningCurvesOffline}
\end{figure}

\begin{figure}[t]
	\centering
	\subfloat[\emph{Cartpole}]{\includegraphics[width=0.5\columnwidth]{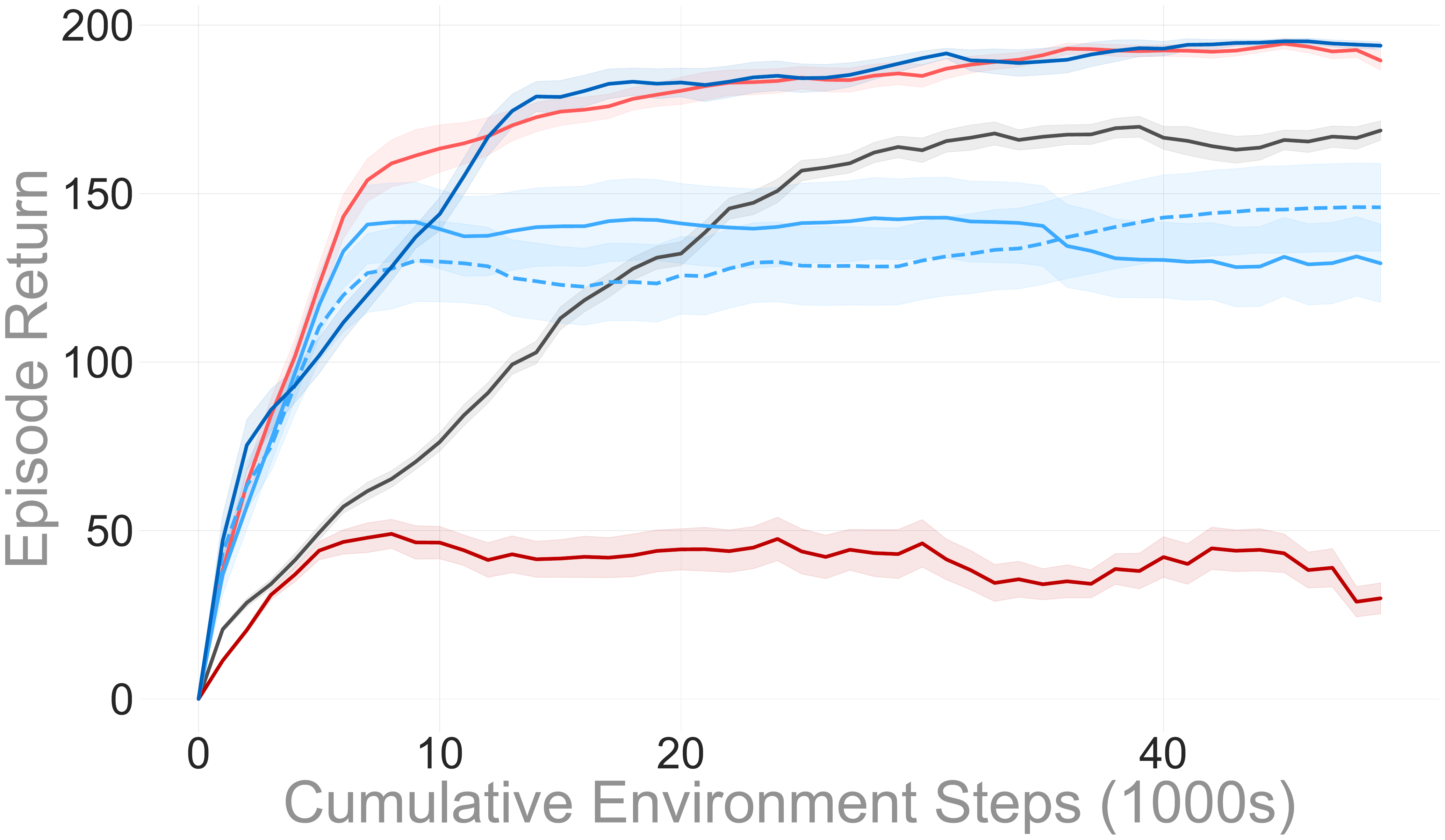}}
	\subfloat[\emph{Catcher}]{\includegraphics[width=0.5\columnwidth]{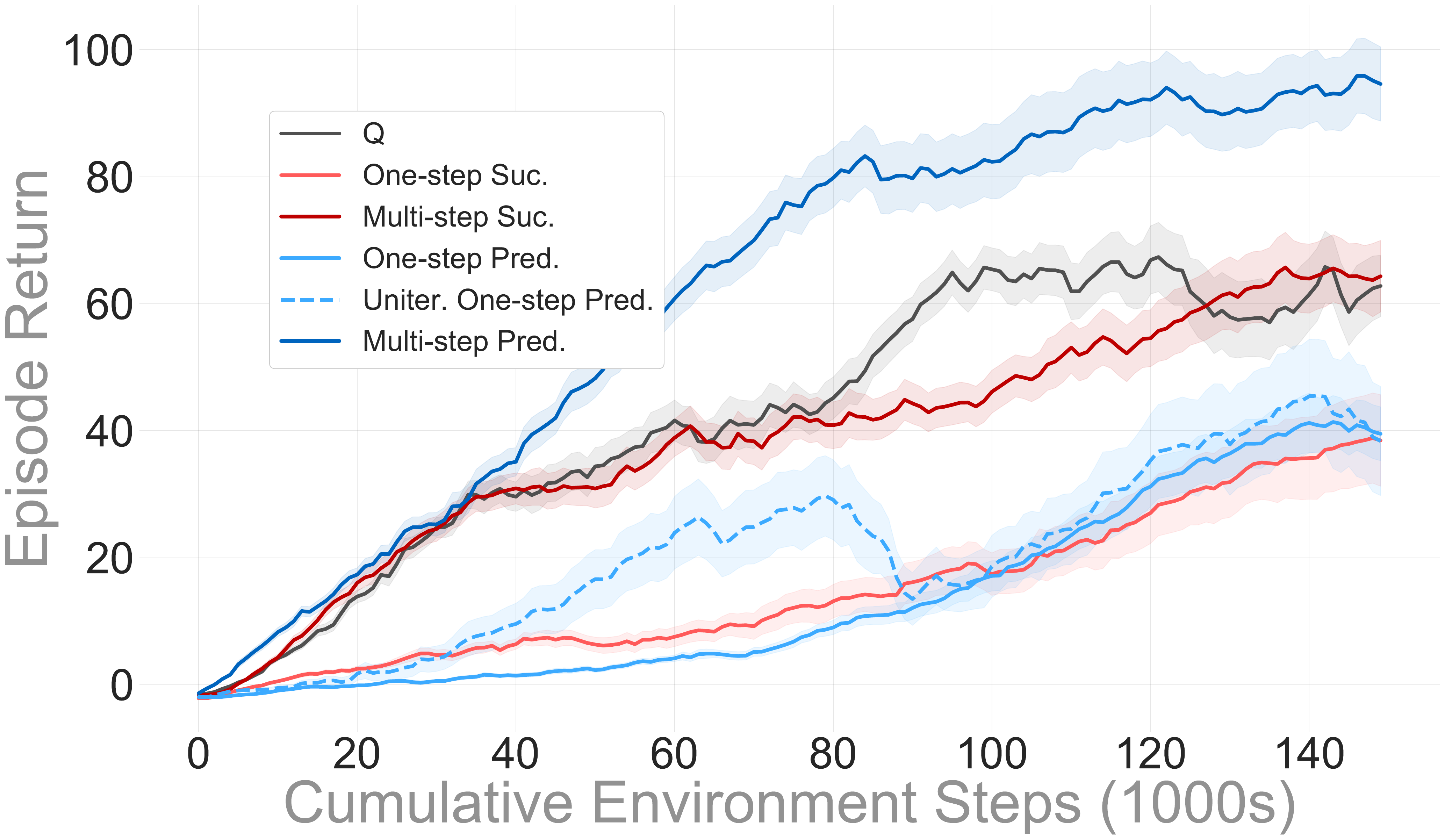}}\\
	\subfloat[\emph{Puddleworld}]{\includegraphics[width=0.5\columnwidth]{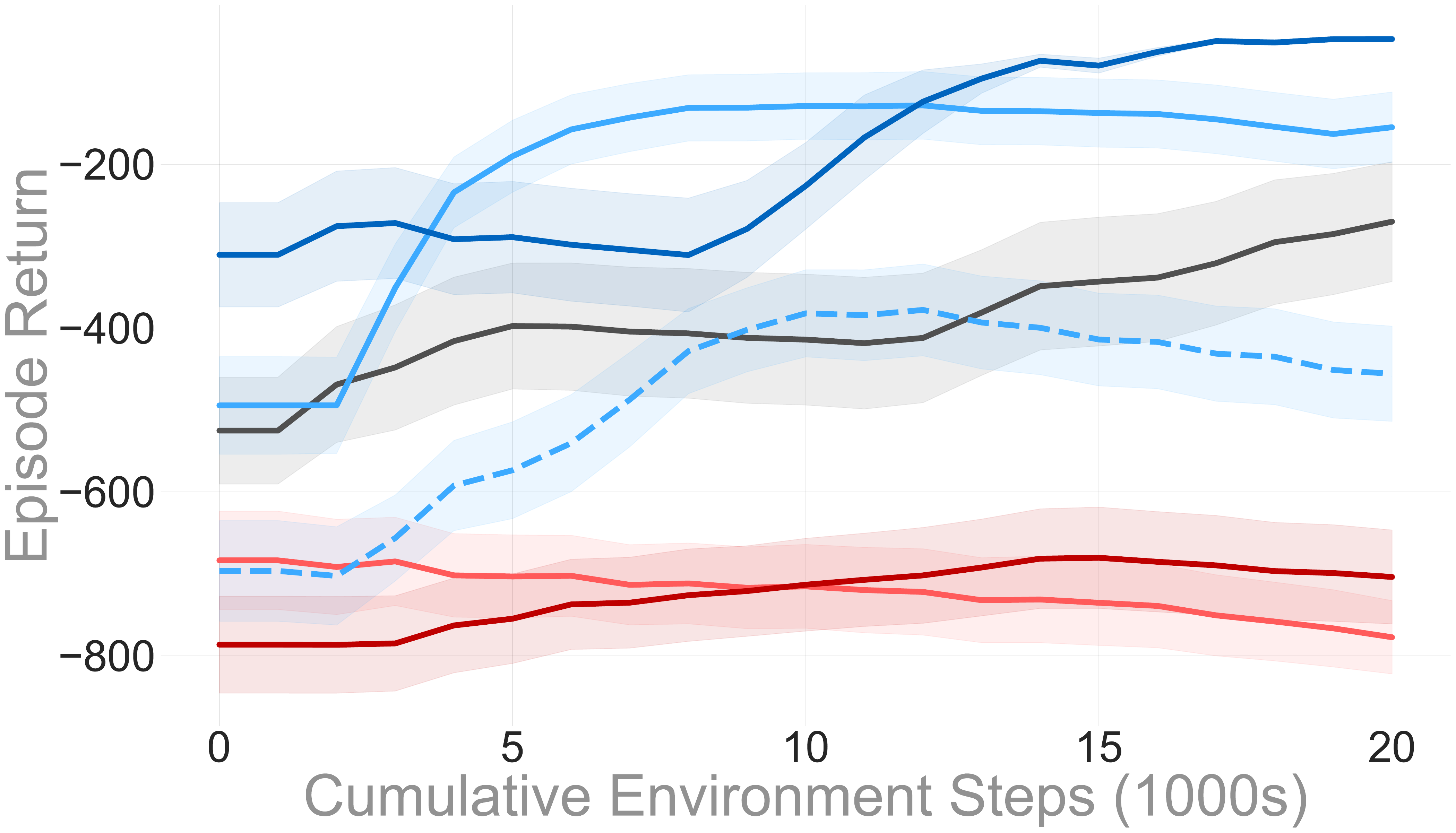}}
	
	\caption{Learning curves for the Dyna variants while updating the environment model \textbf{online}. All curves are averaged over 30 runs, with shaded regions corresponding to standard errors. Multi-step Predecessor performs better in both Catcher and Puddleworld. If an environment model is less prone to HVH, the performance of all variants gets closer to each other, e.g. Cartpole. In online learning, since the model is updated by seeing more samples, the performance of One-step methods might end up being higher up since they perform more planning updates in the background.}
	\label{Figure:Section6/LearningCurvesOnline}
\end{figure}

\subsection{Results when Learning Environment Models Offline} 

Figure \ref{Figure:Section6/LearningCurvesOffline} shows learning curves on the three environments, for a model learned offline. Multi-step Predecessor performs significantly better than the rest of the variants. One-step Successor,  Multi-step Successor, and One-step Predecessor struggle to learn in Catcher and Puddleworld, all performing worse than Q-learning. Each of these algorithms uses a Q-learning update, supplemented with planning updates; the fact that they perform worse than Q-learning indicates that these planning updates are actually harming performance. 

In Cartpole, all the methods except for Multi-step Successor outperform Q-learning. One possible explanation for this difference from the other environments is the structure of the Cartpole environment. The agent starts with the pole balanced, and then is prone to drop it until it learns a near-optimal policy. Multi-step Predecessor might skew the values of these important initial states, by bootstrapping off a value iterated several steps into the future. One-step Predecessor, on the other hand, will reinforce the positive values for these initial states, because it only uses one-step updates, and so updates based on a state near the initial state that still has the pole balanced. On the iterated trajectory, for the states further from this good initial state, it simply decreases already low-valued states, but does not skew the values of the good initial states. Later, we examine the impact of $\beta$ on performance, and find that small $\beta$ are effective in Cartpole, suggesting that one-step updates near real experience are already providing sufficient benefit in this environment. 

Now let us consider the algorithms that do not update real states to simulated states: Multi-step Predecessor Dyna and Uniterated One-step Predecessor Dyna. They both perform better than the other methods on Cartpole and Catcher, but Uniterated One-step Predecessor Dyna performs poorly on Puddleworld, worse than Q-learning. Puddleworld is a gridworld, which is a classic setting where propagating values back quickly is important for sample efficient learning. Removing iteration likely slowed learning enough, that the uniterated variant was unable to learn as quickly as Q-learning. 

\subsection{Results when Learning Environment Models Online} \label{sec_online}

Now we consider if this phenomenon persists if we allow the model to update online. We might expect online updating to resolve the issue because if a high hallucinated value caused an agent to spend a great deal of time in one part of the environment, the model trained \emph{online} on data from these states would eventually become accurate enough to stop predicting these invalid states that have high hallucinated value. This is a possibility. 

However, there are reasons to believe that even if HVH is reduced, it will still have an effect. Correcting the model may take a long time and the agent will make poor choices during this period. Moreover, suppose the model does indeed become more accurate for some regions of states. It is likely that it will become less accurate in other parts of the state space and generate problematic states there. On the other hand, note that online learning only helps if the agent is chasing high hallucinated values. Hallucinated low values can be propagated by planning to make you avoid certain state actions. Not taking those state actions means the model does not see those examples and so cannot fix its hallucinations. 

We show learning curves of three benchmarks in Figure \ref{Figure:Section6/LearningCurvesOnline}, where we updated the environment model every $32$ steps during training. Some outcomes are similar to the offline setting. Of the four Dyna variants, the three that update towards simulated states perform notably worse than Multi-step Predecessor Dyna on all benchmarks.
In Puddleworld, the results are generally very similar, except that updating the models helped the Predecessor algorithms slightly. Otherwise, online model updating generally helped Successor variants more than the Predecessor, and actually causes some of the Predecessor approaches to perform much worse.

Overall, though some specific outcomes changed, the overall trend still indicates that we see similar failures as in the case with offline models. The methods that updated towards simulated states generally performed worse, not providing any benefits over Q-learning. An important next step is to design better model updating approaches, for both forward and backward models, to more fully understand the role of HVH in the online setting. 

\section{The Impact of Model Iteration}\label{Section:beta}
Multi-step Predecessor and Uniterated One-step Predecessor are robust to hallucinated values. Which algorithm is preferred? Here, we focus on $\beta$, a parameter controlling the length of planning rollouts. As described in Section \ref{Section:ADynaDesignSpaceOfDynaAlgorithms}, $\beta$ decays the priority of trajectories commensurate to how many times the model has been iterated to produce a given trajectory. Low values of $\beta$ aggressively decay priority and thus likely result in short rollouts while high values result in priority being primarily a function of TD error. 
High values of $\beta$ are beneficial as they allow for longer rollouts thereby improving sample complexity due to diversity in simulated experience \citep{Holland}. On the other hand, since our imperfect models are susceptible to compounding error, with sufficient iteration the model's predictions may become poor and harmful to learning \cite{Talvitie2014,Talvitie2017}. Ideally, we would like to do some iteration so as to obtain the benefits of diverse planning experience without completely compromising the signal in the model's prediction. This can be achieved by setting $\beta$ to some intermediate value greater than $0$ (no iteration) but less than $1$ (full iteration).

\begin{figure}
	\centering
	\includegraphics[width=0.6\linewidth]{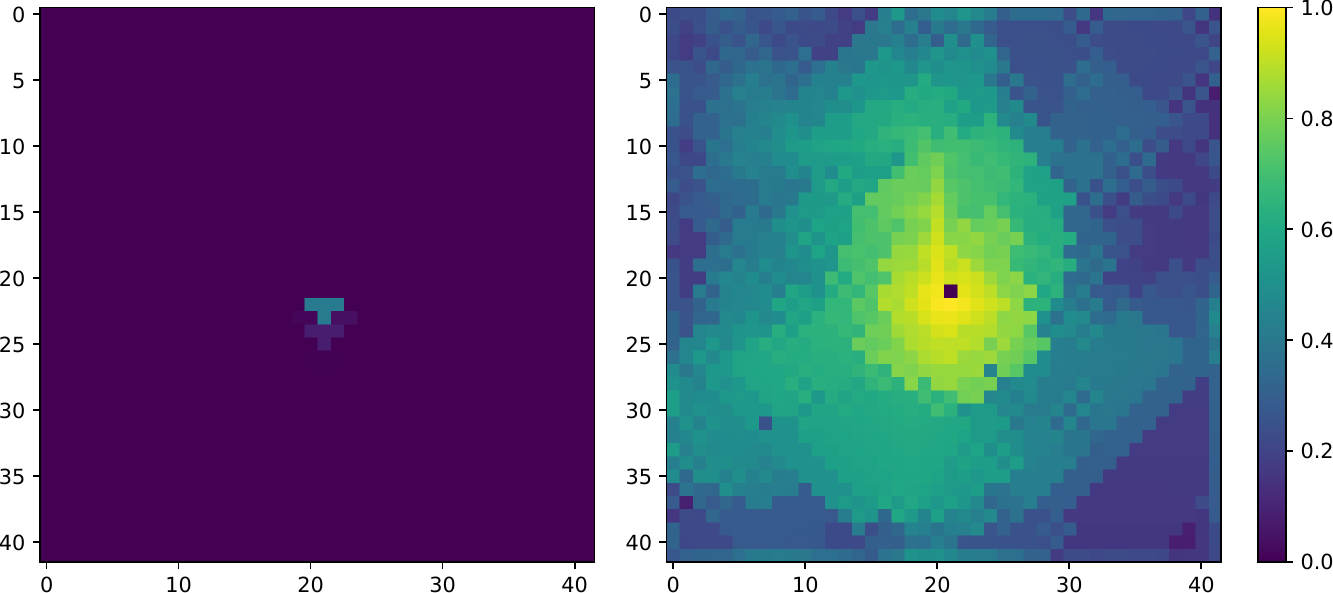}
	\caption{Plot of $\max_a Q(s, a) \:\forall s \in \mathcal{S}$ for $\beta = 0$ (left) and $\beta > 0$ (right) after $2{,}000$ steps.}
	\label{Figure:Section6/Heatmaps}    
\end{figure}

Based on this reasoning, we should expect Multi-step Predecessor Dyna (particularly with a well-tuned $\beta > 0$) to outperform Uniterated One-step Predecessor, which uses a $\beta = 0$.  We exactly saw this outcome in Figure \ref{Figure:Section5/BW_Learning_Curves}, but here we dive a bit deeper. Intuition for
why this might be is seen in Figure \ref{Figure:Section6/Heatmaps},
which shows heatmaps of $\max_a Q(s, a) \:\forall s \in \mathcal{S}$
on a non-optimistically initialised agent on Borderworld. Multi-step Predecessor is much more efficient at propagating value
information. Unlike Uniterated One-step Predecessor, which generates a
single predecessor, updates it and then discards the trajectory, with Multi-step Predecessor, we can generate trajectories radiating
backwards from a particular state.

\begin{figure}
	\centering
	\begin{subfigure}{.45\textwidth}
	    \begin{subfigure}{\textwidth}
          \renewcommand\thesubfigure{\alph{subfigure}1}
          \centering
          \includegraphics[width=0.95\textwidth, height=0.2\textheight]{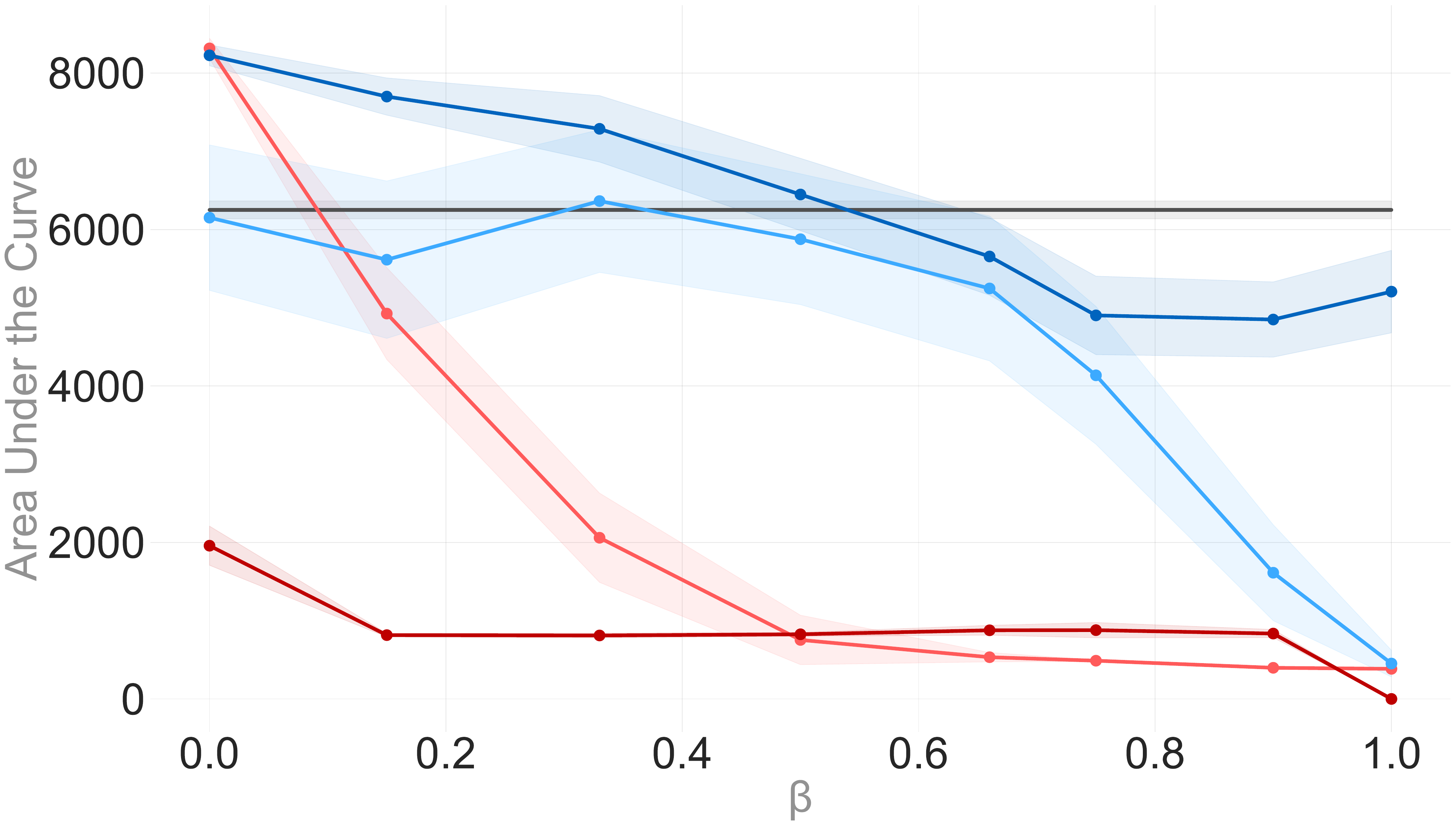}
          \caption{\emph{Cartpole-Online}}
          \label{cartpole_learning_curve_online}
        \end{subfigure}
        \begin{subfigure}{\textwidth}
          \addtocounter{subfigure}{-1}
          \renewcommand\thesubfigure{\alph{subfigure}2}
          \centering
          \includegraphics[width=0.95\textwidth, height=0.2\textheight]{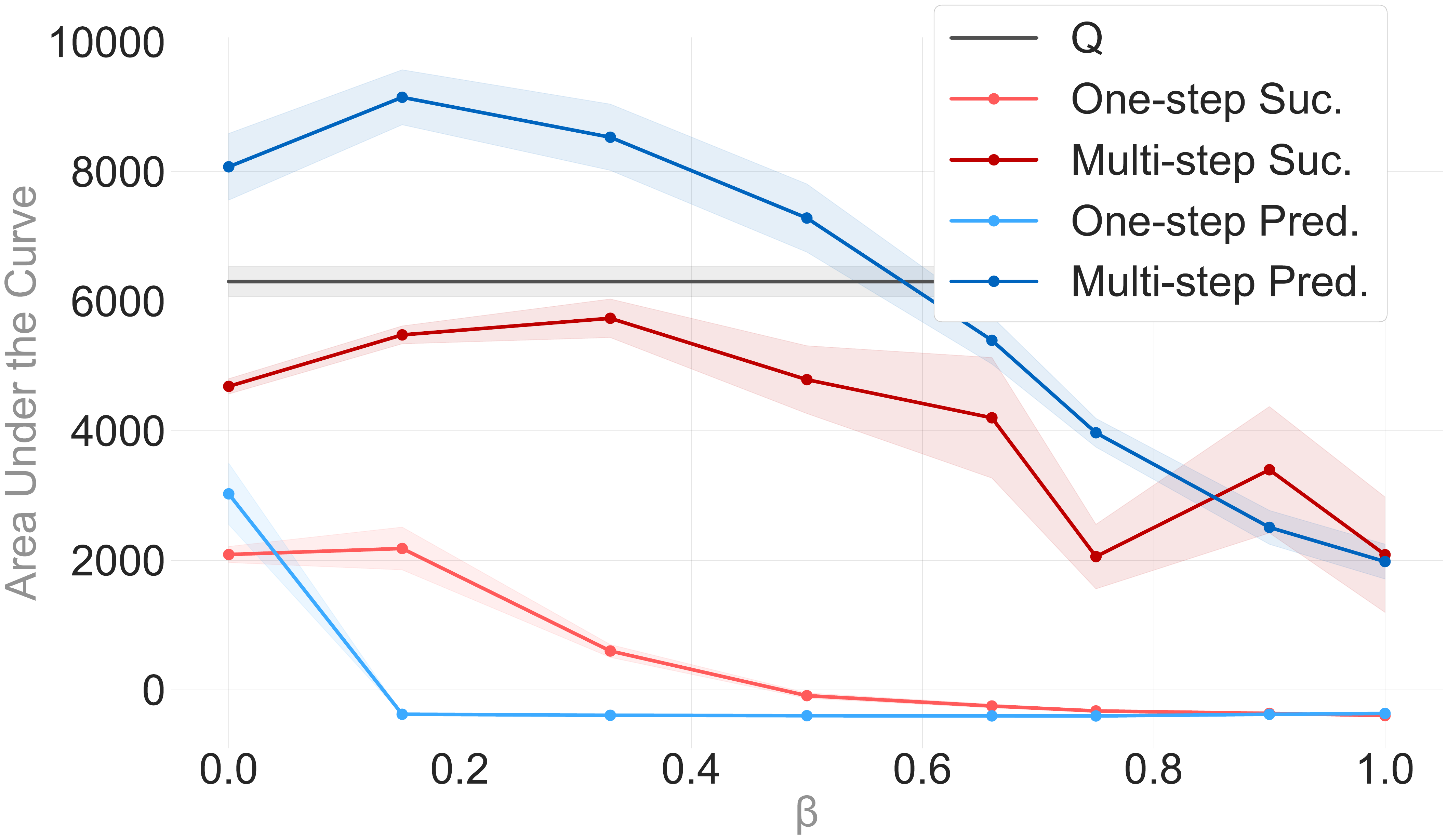}
          \caption{\emph{Catcher-Online}}
          \label{catcher_learning_curve-online}
        \end{subfigure}
        \begin{subfigure}{\textwidth}
          \addtocounter{subfigure}{-1}
          \renewcommand\thesubfigure{\alph{subfigure}3}
          \centering
          \includegraphics[width=0.95\textwidth, height=0.2\textheight]{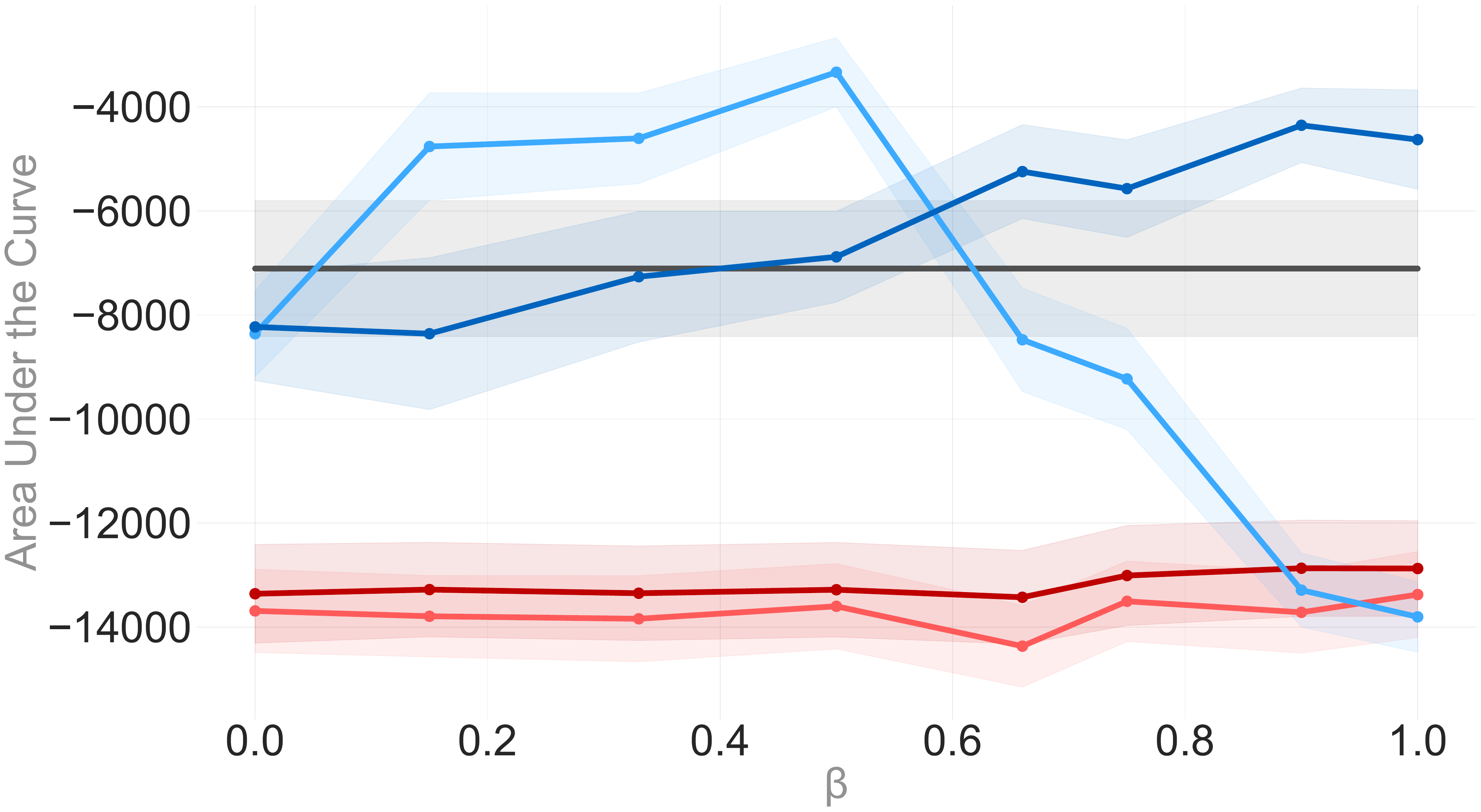}
          \caption{\emph{Puddleworld-Online}}
          \label{puddleworld_learning_curve_onlin}
        \end{subfigure}
        \addtocounter{subfigure}{-1}
		\caption{Updating environment model online}
		\label{Online_update_model}
	\end{subfigure}
	\begin{subfigure}{.45\textwidth}
	    \begin{subfigure}{\textwidth}
          \renewcommand\thesubfigure{\alph{subfigure}1}
          \centering
          \includegraphics[width=0.95\textwidth, height=0.2\textheight]{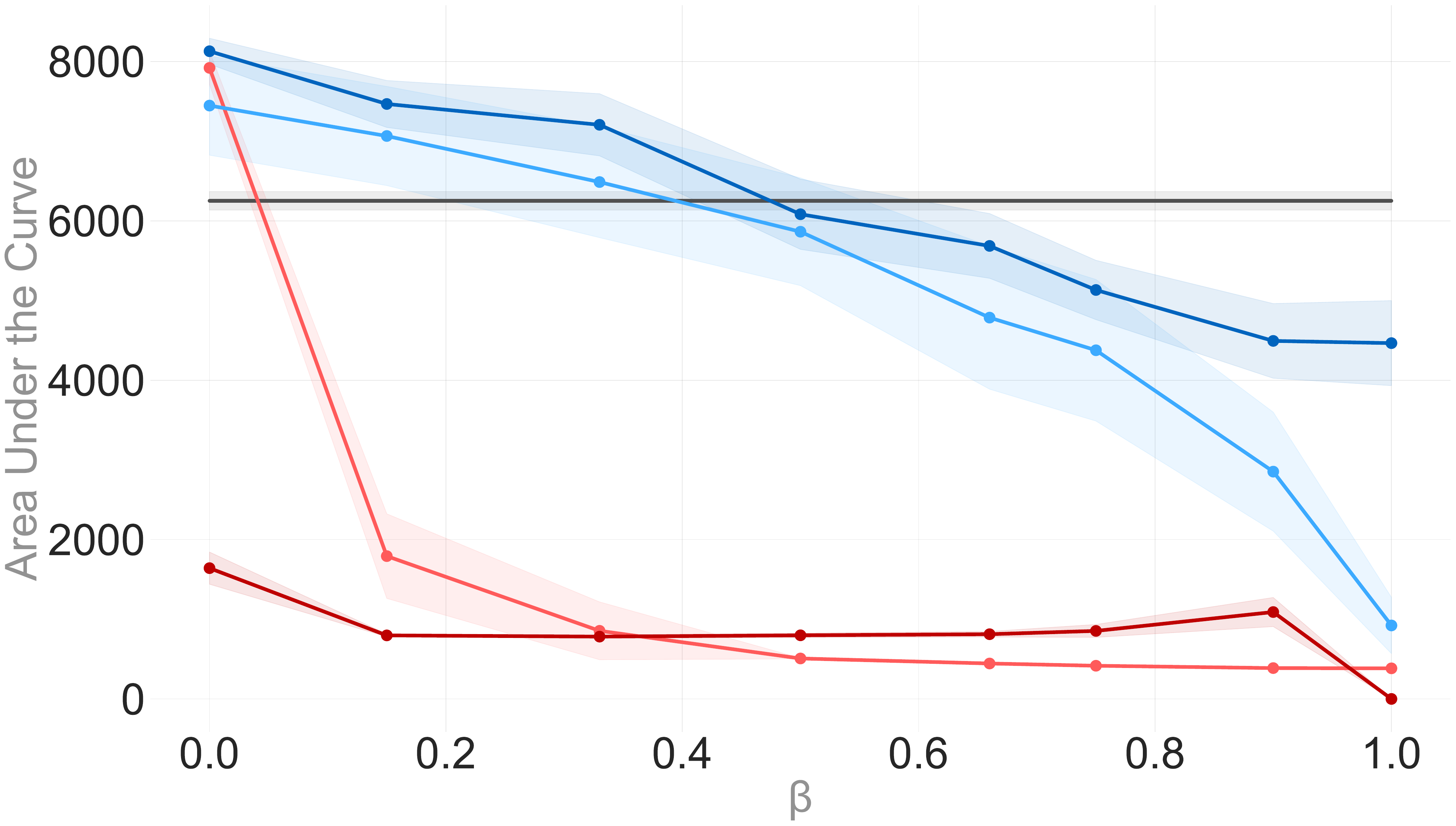}
          \caption{\emph{Cartpole-Offline}}
          \label{cartpole_learning_curve-offline}
        \end{subfigure}
        \begin{subfigure}{\textwidth}
          \addtocounter{subfigure}{-1}
          \renewcommand\thesubfigure{\alph{subfigure}2}
          \centering
          \includegraphics[width=0.95\textwidth, height=0.2\textheight]{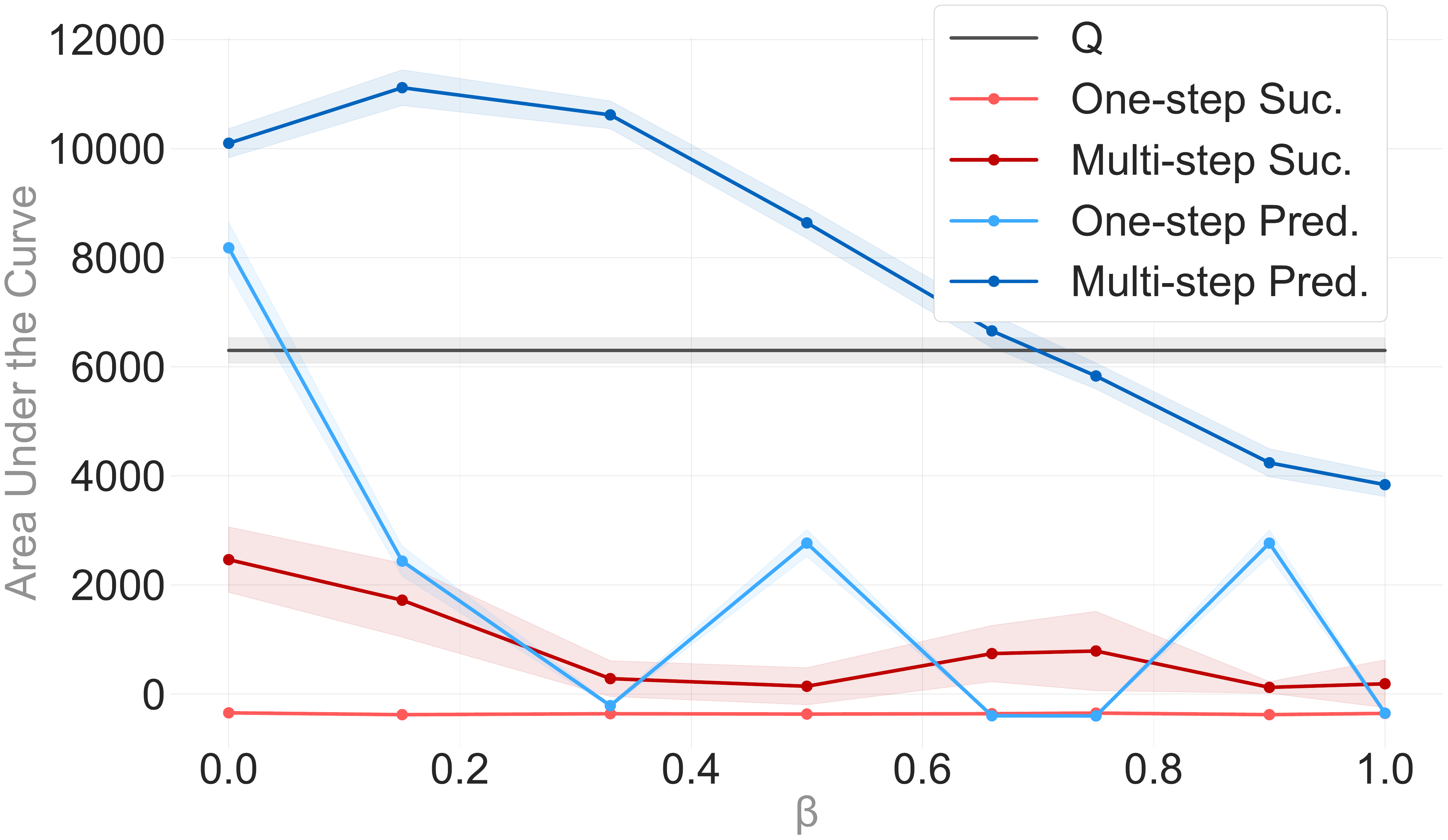}
          \caption{\emph{Catcher-Offline}}
          \label{catcher_learning_curve-offline}
        \end{subfigure}
        \begin{subfigure}{\textwidth}
          \addtocounter{subfigure}{-1}
          \renewcommand\thesubfigure{\alph{subfigure}3}
          \centering
          \includegraphics[width=0.95\textwidth, height=0.2\textheight]{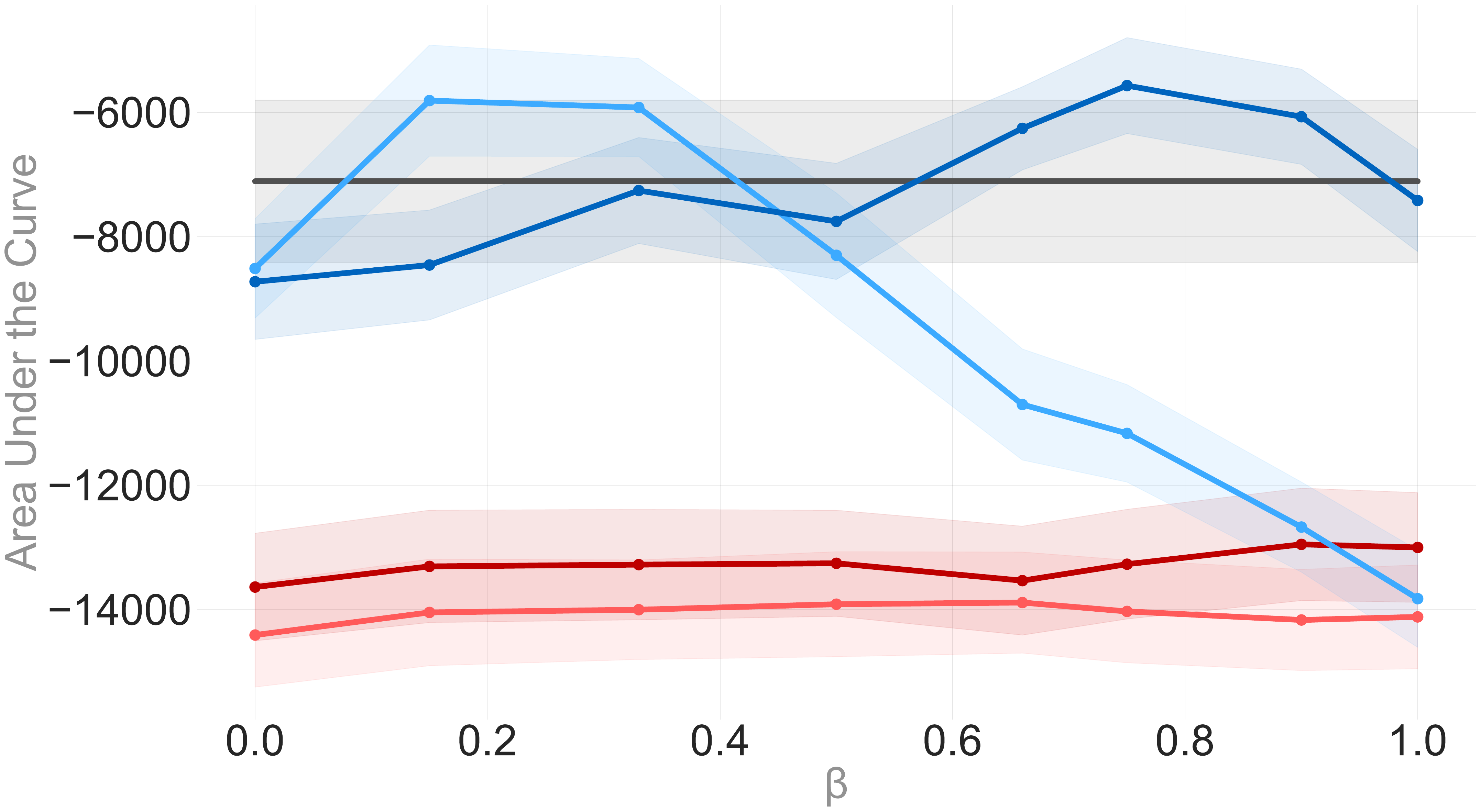}
          \caption{\emph{Puddleworld-Offline}}
          \label{puddleworld_learning_curve_offlin}
        \end{subfigure}
        \addtocounter{subfigure}{-1}
		\caption{Use pre-trained environment model}
		\label{Offline_model}
	\end{subfigure}
	\caption[\textbf{(a):} []{ Performance versus $\beta$, where $\beta$ controls how far back the model is iterated. The priorities are multiplied by $\beta^n$, making it more and more unlikely to select the trajectory and expand it further. Multi-step Predecessor is generally the best-performing algorithm, across $\beta$. Note that One-step Predecessor with $\beta=0$ is Uniterated One-step Predecessor.
	}
	\label{Figure:Section6/Beta_Sensitivity}
\end{figure}

We can also look at the performance versus $\beta$ on the three benchmark domains, both with offline models and models updated online. 
Figure \ref{Figure:Section6/Beta_Sensitivity} plots the AUC versus $\beta$, where the AUC means the area under the learning curve. Uniterated One-step Predecessor corresponds to Multi-step Predecessor at $\beta = 0$. The best performance with offline models for all three environments is
attained by Multi-step Predecessor with some intermediate value of
$\beta$. 

In the online setting, we can actually see that what seemed to be the best algorithms in Section \ref{sec_online} actually only achieved this performance for one specific $\beta$ and otherwise performed poorly. One-step Successor in Cartpole performed well for $\beta = 0$, and very poorly otherwise. This result is somewhat surprising, given Multi-step Successor performed poorly for all $\beta$, including $\beta = 0$. The only difference between the two for $\beta = 0$ (no iteration) is that for a tuple of real experience $(s,a,s',r)$, after generating $(a', s'', r'')$ as, the One-step updates $(s',a')$ towards $r'' + \gamma \max_{a''}Q(s'',a'')$ and Multi-step updates $(s,a)$ towards $r' + \gamma r'' + \gamma^2 \max_{a''}Q(s'',a'')$. It is possible that here, in our limited sweep over hyperparameters, the set of stepsizes was more suitable for One-step Successor for this one specific $\beta$ value.

\section{Conclusion and Discussion}\label{Section:Conclusion}

We introduced Multi-step Predecessor Dyna: a Dyna-style algorithm that uses a backward model and multi-step updates to avoid compounding errors in planning. We motivate our proposed approach by investigating one potentially problematic type of model error for Dyna: \emph{hallucinated values}. 
We present the \emph{Hallucinated Value Hypothesis} (HVH): updates to values of real states towards values of simulated states result in misleading action values which adversely affect the control policy. 

More specifically, we discussed four Dyna variants based on using forward or backward environment dynamics models, and one-step versus multi-step temporal difference (TD) updates. We show three of these---which correspond to widely used Dyna algorithms---are prone to failure due to the HVH. This is because some simulated states could correspond to hallucinated states---unreachable or non-existent states---with arbitrary values that are not updated with real experience. Updates that bootstrap off of such states may propagate this arbitrary value.
However, the fourth variant---our new algorithm Multi-step Predecessor Dyna---cannot suffer from the HVH by design: it only updates towards values of real states.
We provide evidence for the HVH by demonstrating that the three former variants fail whereas the latter does not in a toy domain and three standard benchmark environments. 
We also highlighted that iteration is worthwhile, by comparing to an uniterated variant that also should not suffer from the HVH. 
Multi-step Predecessor Dyna allows one to gain the benefits of diverse experience offered by iterating a model without succumbing to model error. This work provides some practical tools to more carefully incorporate model iteration into Dyna.

One limitation in using backwards models is that we might end up having a non-deterministic model instead of a deterministic one in a backwards setting. Any pair of (previous action-state), $(a_{t-1}, s_t)$, might have more than only one predecessor state. In a backwards model, we may not know exactly which one of the predecessor states we came from. There is some other research in this regard \cite{edwards2018forwardbackward} where they keep a parallel forward model to keep track of the deterministic trajectories. \cite{chelu2020forethought} provide several new objectives to improve learning backwards models. Advances in estimating backwards models will make Multi-step Predecessor Dyna an even more practical avenue to pursue.

\acks{We would like to thank our generous funders for supporting this work, specifically the NSERC Discovery grant program and CIFAR for funding the Alberta Machine Intelligence Institute and the CIFAR Canada AI Chairs program. 
}

\appendix 
\section{A Discussion on Trajectory Screening Approaches} \label{appendix A}

Let us return to the question about correcting off-policy trajectories. We considered the following screening strategies for trajectories from the priority queue.
\begin{enumerate}
    \item \textbf{[No Screening]} The most naive strategy is to update action-values for all trajectories and actions, which is the most biased approach because the trajectories are likely to be off-policy. We discard off-policy trajectories after updating to prevent further biased planning and updates on them.
    \item \textbf{[Conservative Screening]} We only update the action values if the whole trajectory is on-policy, otherwise, we just drop the trajectory. This conservative approach might lead to an empty priority queue and no planning updates. Consequently, the multi-step approaches would get fewer planning updates, since it is less likely to have a fully on-policy multi-step trajectory. 
    \item \textbf{[On-policy Subchunks]} To pick a middle ground---between dropping the whole trajectory or complete ignoring the bias---we can use an on-policy sub-chunk. In other words, for the backwards model, we start from the current state $s_t$ and go backwards in time $s_{t-i}, i= 1,2,3 ... $ along the trajectory until we find the first off-policy action. We drop the rest of the trajectory (keeping the off-policy action) and update with this on-policy sub-chunk. Similarly, in forward models, we go forward in time $s_{t+i}, i = 1, 2, 3, \cdots$ and only keep the on-policy portion.          
\end{enumerate}
As a small nuance on the last point, because we are learning action-values, the first action for a return need not be on-policy, because we consider the return given a state and action. For the backwards model, this means we keep that first off-policy action, since we will update from $Q(s_{t-i}, a_{t-i})$ towards an on-policy return. For the forwards model, the very first action can be off-policy, but the remaining must be on-policy. 
Algorithms \ref{Algorithm:Screening} and \ref{Algorithm:is_on_policy} show the details of the screening approaches.


\begin{algorithm}[t]
\caption{Pop Tuple and Screen}
\label{Algorithm:Screening}
\begin{algorithmic}[1]
    \State \textbf{Input} Priority queue $P$
    \State $expanding \gets True$
        \If{\texttt{No\_Screening}}
            \State Pop tuple $T = (s_t, a_t, ..., s_{t+n}, n)$ from $P$ 
            \If{$\texttt{NOT } IsOnPolicy(T_n)$}
                \State $expanding \gets False$
            \EndIf
            \State $T'_{n} \gets T_{n}$
        \ElsIf{\texttt{Conservative\_Screening}}
            \While{$P \texttt{ not empty}$}
                \State Pop tuple $T_n = (s_t, a_t, ..., s_{t+n}, n)$ from $P$ 
                \If{$IsOnPolicy(T_n)$}
                    \State $T'_n \gets T_n$
                    \State $Break$
                \EndIf
                
            \EndWhile
        \ElsIf{\texttt{On\_policy\_subchunks}}
            \State Pop tuple $T = (s_t, a_t, ..., s_{t+n}, n)$ from $P$ 
            \If{Forward}
                \State $s_{t}, a_{t}, r_{t}, s_{t+1} \gets get\_first\_tuple(T_n)$
                \For{$i=1 \textit{ to } n$}
                    \If{$a_{t+i} \neq \argmax_{a'} Q(s_{t+i}, a')$}
                        \State $T_n \gets (s_{t}, a_{t}, r_{t+1}, ..., s_{t+i}, i)$ 
                        \Comment{Last action could be off-policy.}
                        \If{$\texttt{NOT } IsOnPolicy(T_n)$}
                            \State $expanding \gets False$
                        \EndIf
                    \EndIf
                \EndFor                
            \EndIf
            \If{Backward}
            \State $s_{t+n-1}, a_{t+n-1}, r_{t+n}, s_{t+n} \gets get\_last\_tuple(T_n)$
                \For{$i=n-1 \textit{ to } 0$}
                    \If{$a_{t+i} \neq \argmax_{a'} Q(s_{t+i}, a')$}
                        \State $T_n \gets (s_{t+i}, a_{t+i} ..., r_{t+n}, s_{t+n}, n-i)$
                        \Comment{First action could be off-policy.}
                    \EndIf
                \EndFor
                \If{$\texttt{NOT } IsOnPolicy(T_n)$}
                    \State $expanding \gets False$
                \EndIf
            \EndIf
            \State $T'_n \gets T_n$
        \EndIf    
    \Return $T'_{n}, \texttt{ expanding}$
\end{algorithmic}
\end{algorithm}

\begin{algorithm}[t]
\caption{IsOnPolicy}
\label{Algorithm:is_on_policy}
\begin{algorithmic}[1]
    \State \textbf{Input} Trajectory $T_n = (s_{t}, a_{t}, r_{t+1}, ..., s_{t+n}, n)$
    \For{$i=0 \textit{ to } n$}
        \State $s_{t+i}, a_{t+i}, r_{t+i}, s_{t+i+1} \gets get\_tuple(T_n, i)$
        \If{$a_{t+i} \neq Argmax_{a'} Q(s_{t+i}, a')$}
            \State \Return $False$
        \EndIf
    \EndFor

\State \Return $True$

\end{algorithmic}
\end{algorithm}

\begin{figure}[t]
	\centering
	\subfloat[\emph{Conservative, Fewer Updates}]{\includegraphics[width=0.5\linewidth]{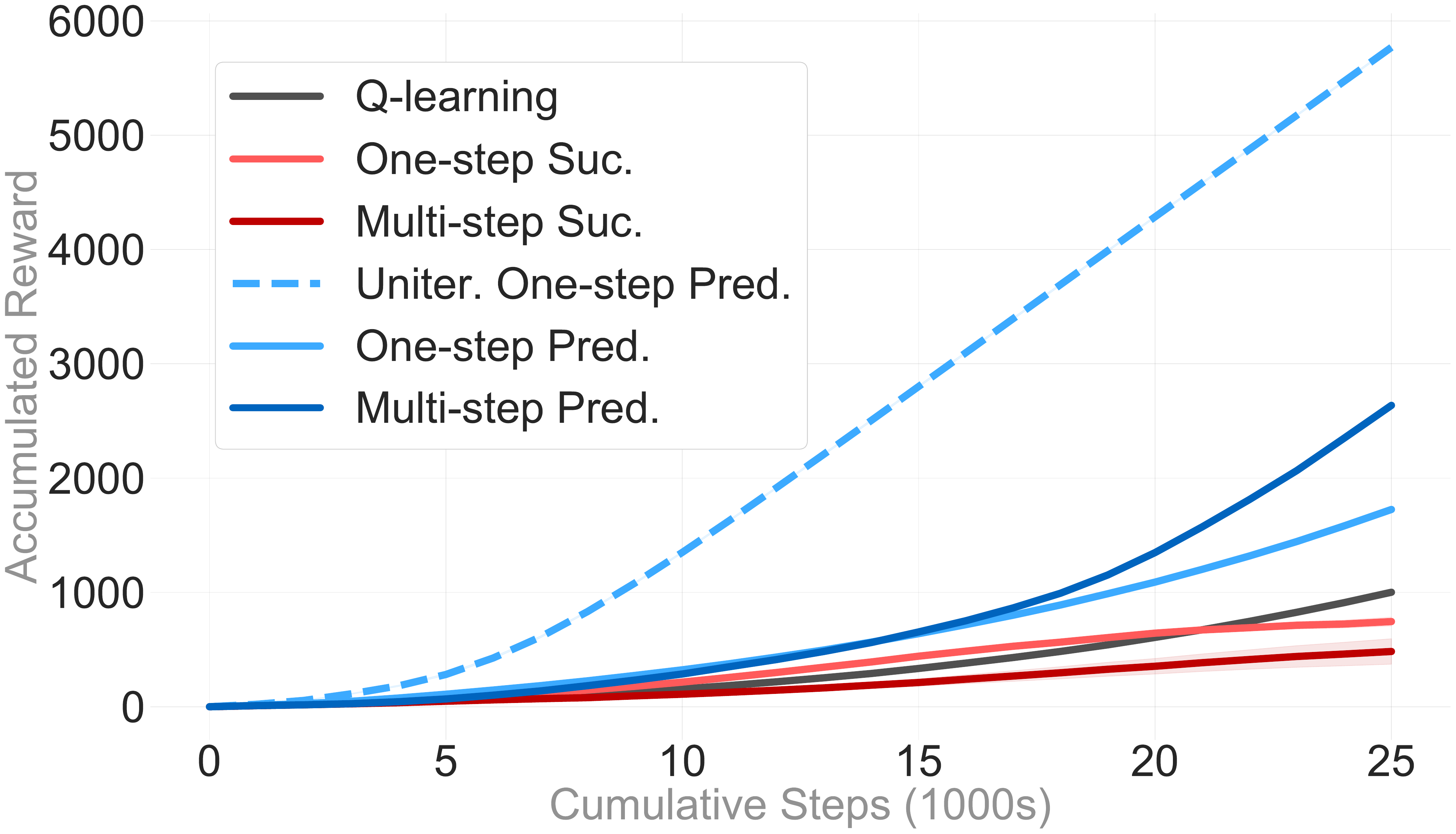}}
	\subfloat[\emph{Conservative, More Updates}]{\includegraphics[width=0.5\linewidth]{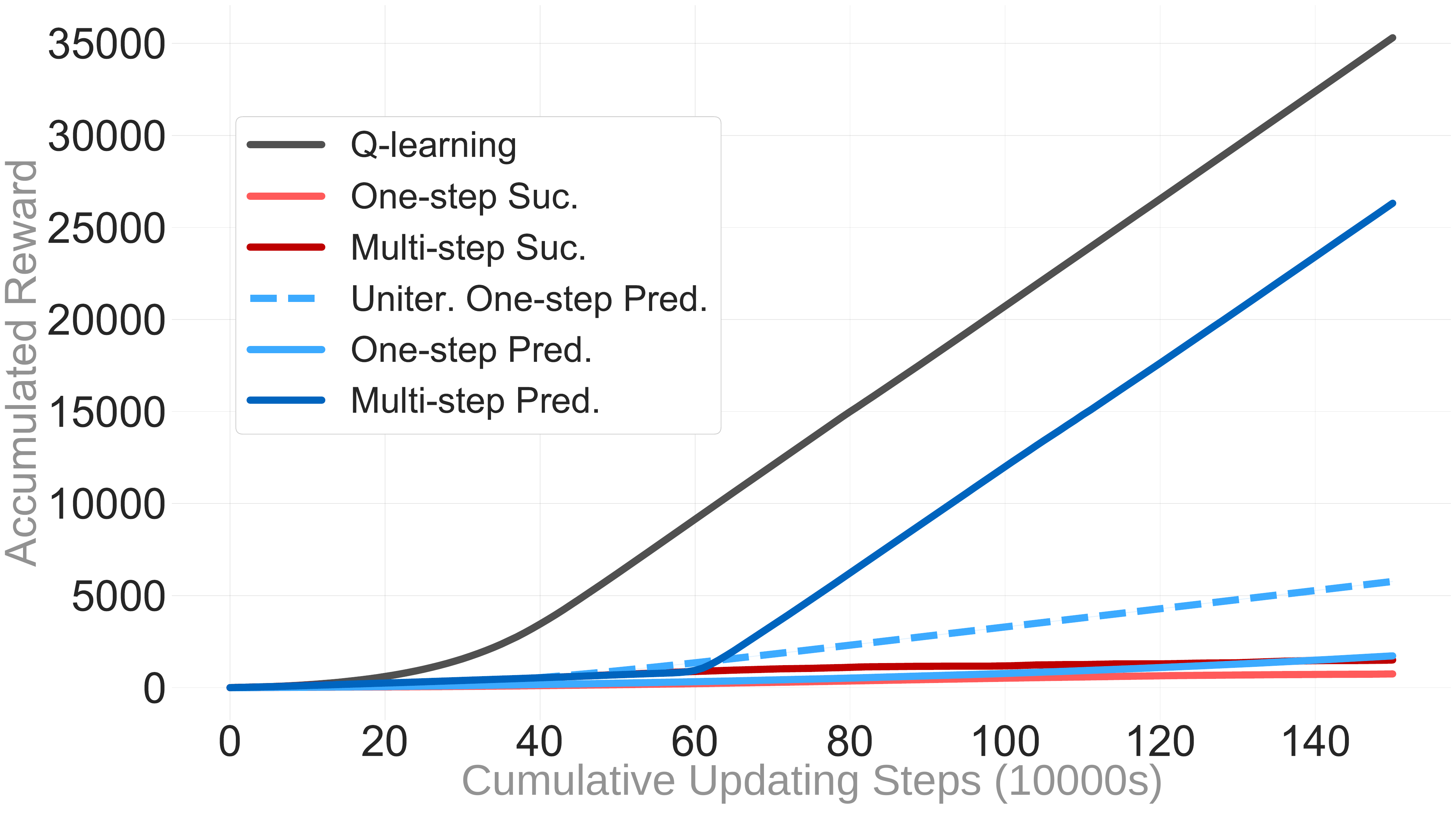}}\\
	\subfloat[\emph{No Screening, Drop Off-policy Trajectories}]{\includegraphics[width=0.5\linewidth]{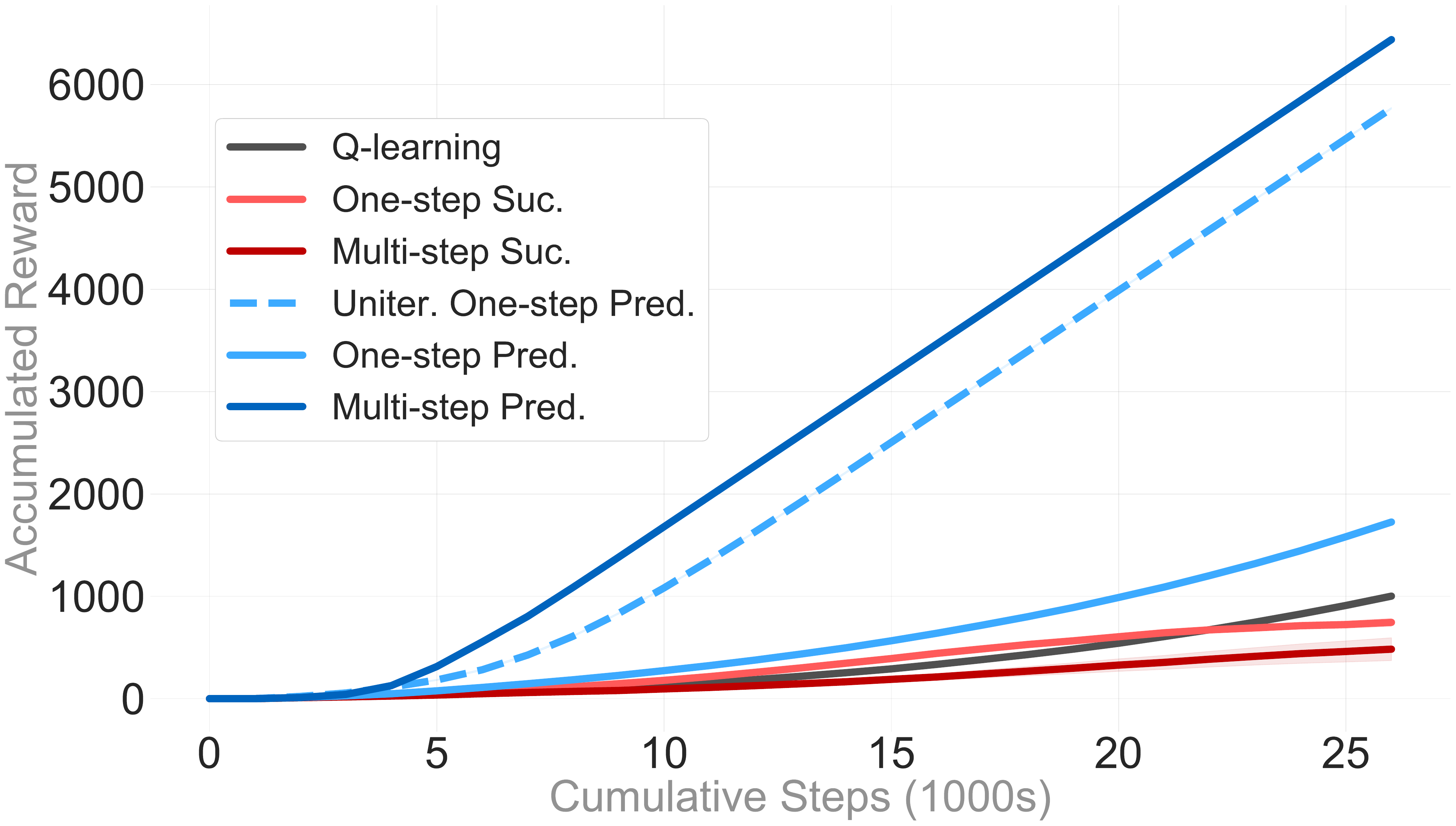}}
 \subfloat[\emph{On-policy Subtrajectories (Approach in Fig. \ref{Figure:Section5/BW_Learning_Curves})}]{\includegraphics[width=0.5\linewidth]{./Figures/Section5/BW_Learning_Curves_Screening_OffPolicy_SubChunks_CumSteps.pdf}}
 	\caption{Comparing different multi-step screening approaches. Error bars are not visible as they are smaller than line thicknesses.}
	\label{Figure:screening-approaches}
\end{figure}

We start by showing results with the conservative approach in Figure \ref{Figure:screening-approaches}-(a), where the Q-learning with a conservative screening approach mimics the result of Expected-Sarsa.
The screening method ignores all off-policy trajectories without any updates during planning in multi-step variants.
All one-step methods, however, perform a planning update regardless of the last transition
of popped out trajectory being off-policy or on-policy.
One-step methods end up with way more updates to their Q-values (5x) rather as compared to the Multi-step methods. The results show that Multi-step Predecessor variant still outperforms the rest of the algorithms. However, Uniterated One-step Predecessor performs better than Multi-step Predecessor Dyna, as it avoids hallucinated values but gets to perform many more updates.
In other words, the trade-off here is about whether to perform a biased update using an off-policy trajectory that might distort Q-value or no update at all. Here, we chose no updates to keep the amount of computation approximately the same otherwise, we could just keep popping until we find a valid trajectory during the planning step. No updates are more similar to what we would do if we did Expected-Sarsa, where we would update with Important Sampling (IS) ratios. Then we would do every update during planning, but many of them would be negligible. This is one of the reasons we chose to stick with Q-learning and screening rather than explicitly implement Expected-Sarsa.



The next empirical choice of screening is to only allow unbiased updates with on-policy trajectories regardless of differences in the computational complexity of algorithms. Therefore, in each planning step, we keep popping out trajectories until either the queue is empty or we find an on-policy trajectory. The drawback of this choice is that some algorithms might see more true sampled data from the environment than the others. Q-learning in these experiments would perform as many updates based on real experience as all other four variants would do either through planning or interaction with the environment. Moreover, Multi-step variants would probably see more real samples rather than One-steps since their queues become empty more frequently before they could find any on-policy trajectories. 
Figure \ref{Figure:screening-approaches}-(b) shows the learning curve based on the accumulated reward as a function of updates for different Dyna variants as well as Q-learning. As we expected, Q-learning performs best since its updates are all based on true samples from agent's real experiences rather than hallucinated ones from planning in the background. Comparing approaches based on the updating steps results in Q-learning having the best performance and least sample efficiency. Even in this extreme choice of evaluation, we can still see that Multi-step Predecessor and Uniterated One-step Predecessor still outperform the rest of Dyna variants due to preventing hallucinated values. Multi-step Predecessor is shown to be as a shifted version of Q-learning with a better sample efficiency since it only applies unbiased TD updates in planning steps and prevents the negative effects of model errors in Q-value updates. These results clearly show the reliability of Multi-step Predecessor Dyna compared to the rest of the common Dyna variants. Note that we are not competing against Q-learning in terms of the number of updates. The whole purpose of Model-Based Reinforcement Learning (MBRL) is to allow more updates in the background. The only reason to control planning steps for MBRL agents is to avoid that being a confounding factor. 

If we simply perform no screening, but also drop the off-policy trajectories after planning updates instead of expanding them and putting them back into the queue, we would get the results depicted in Figure \ref{Figure:screening-approaches}-(c). 

Figure \ref{Figure:screening-approaches}-(d) demonstrates similar learning curves for on-policy subchunk screening approach. The reason is that rolling out the on-policy subchunks only happens for a few trajectories, i.e. the length of the most of the trajectories is less equal then $2$. This means that in both no-screening and on-policy subchunk screening methods, we still update agents after seeing the first off/on-policy state-action pair, e.g. $(s_t, a_t)$, and then in no-screening we drop the next state-action pair (e.g. $(s_{t+1}), a_{t+1}$ in forward and $(s_{t-1}, a_{t_1})$ in backwards), while in on-policy subchunk approach we take $(s_{t+1}), a_{t+1}$ in forward and $(s_{t-1}, a_{t_1})$ in backwards to plan further by rolling out on them. However, in agent's perspective, none of these two screening approaches are much of a difference since agent would probably still see other trajectories starting from $(s_{t+1}), a_{t+1}$ or $(s_{t-1}, a_{t_1})$ with no-screening based on its interactions with the environment any ways.

\begin{figure}[h!]
	\centering
	\subfloat[\emph{Cartpole}]{\includegraphics[width=0.5\columnwidth]{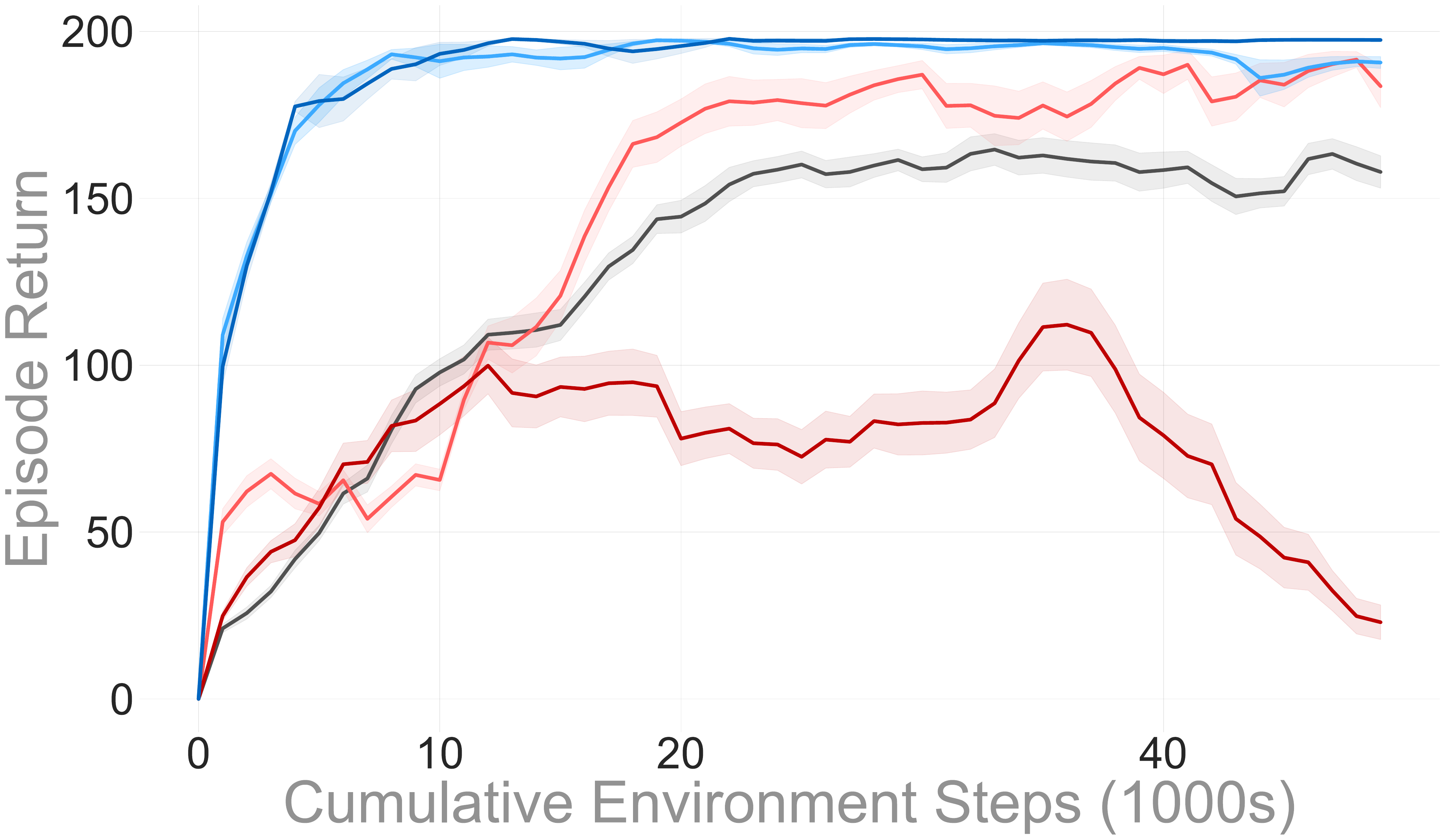}}
	\subfloat[\emph{Catcher}]{\includegraphics[width=0.5\columnwidth]{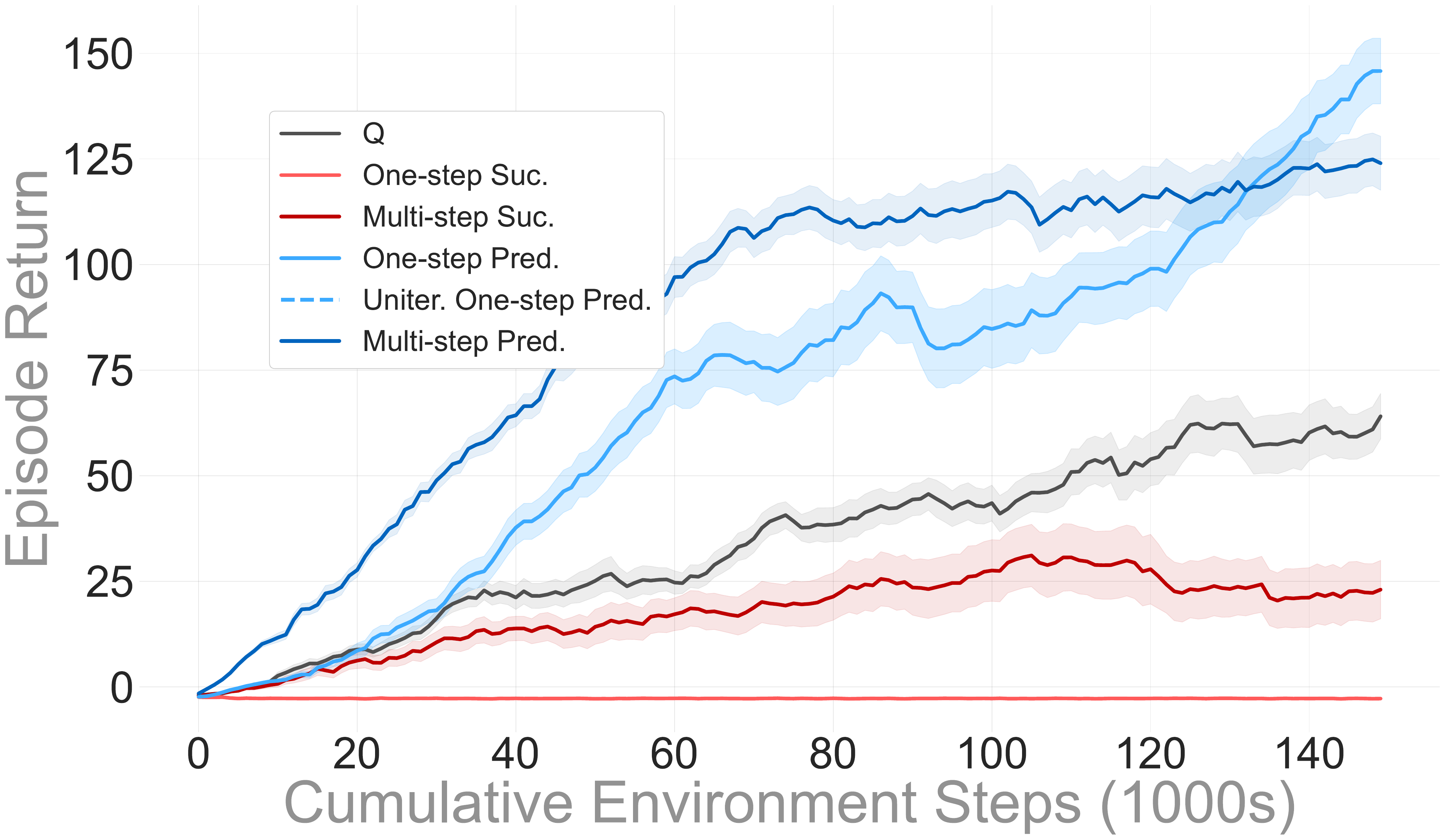}}\\
	\subfloat[\emph{Puddleworld}]{\includegraphics[width=0.5\columnwidth]{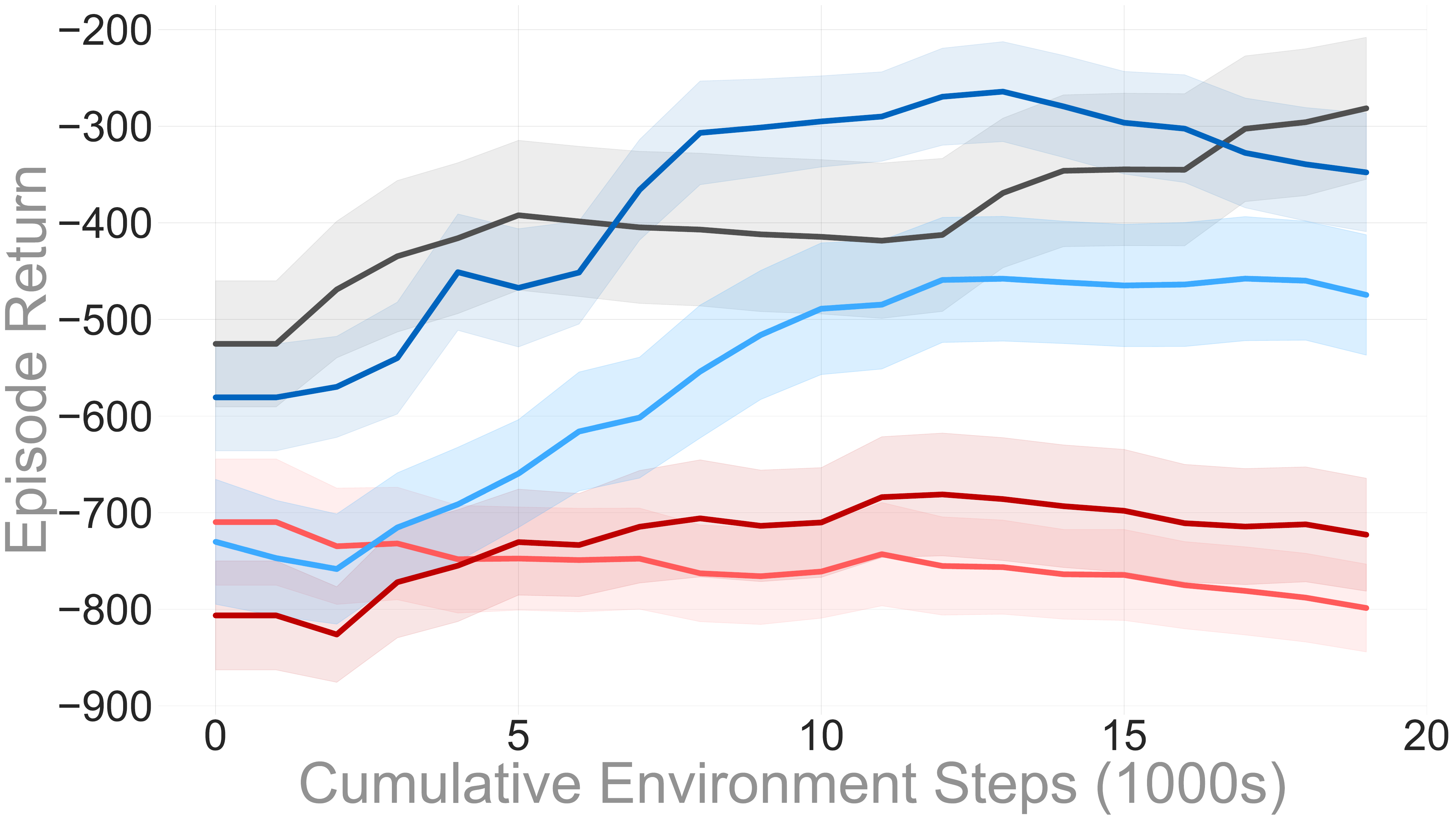}}
	
	\caption{Learning curves for the Dyna variants with a fixed environment model learned \textbf{offline} with a conservative screening approach. All curves are averaged over 30 runs, with shaded regions corresponding to standard errors. The blue lines are predecessor models and the red lines are successor models. We can see generally that Dyna with predecessor models is more effective, and that the variants that Multi-step Predecessor performs significantly better in all environments.}
	\label{Figure:Section6/LearningCurvesOfflineconservative}
\end{figure}

\begin{figure}[h!]
	\centering
	\subfloat[\emph{Cartpole}]{\includegraphics[width=0.5\columnwidth]{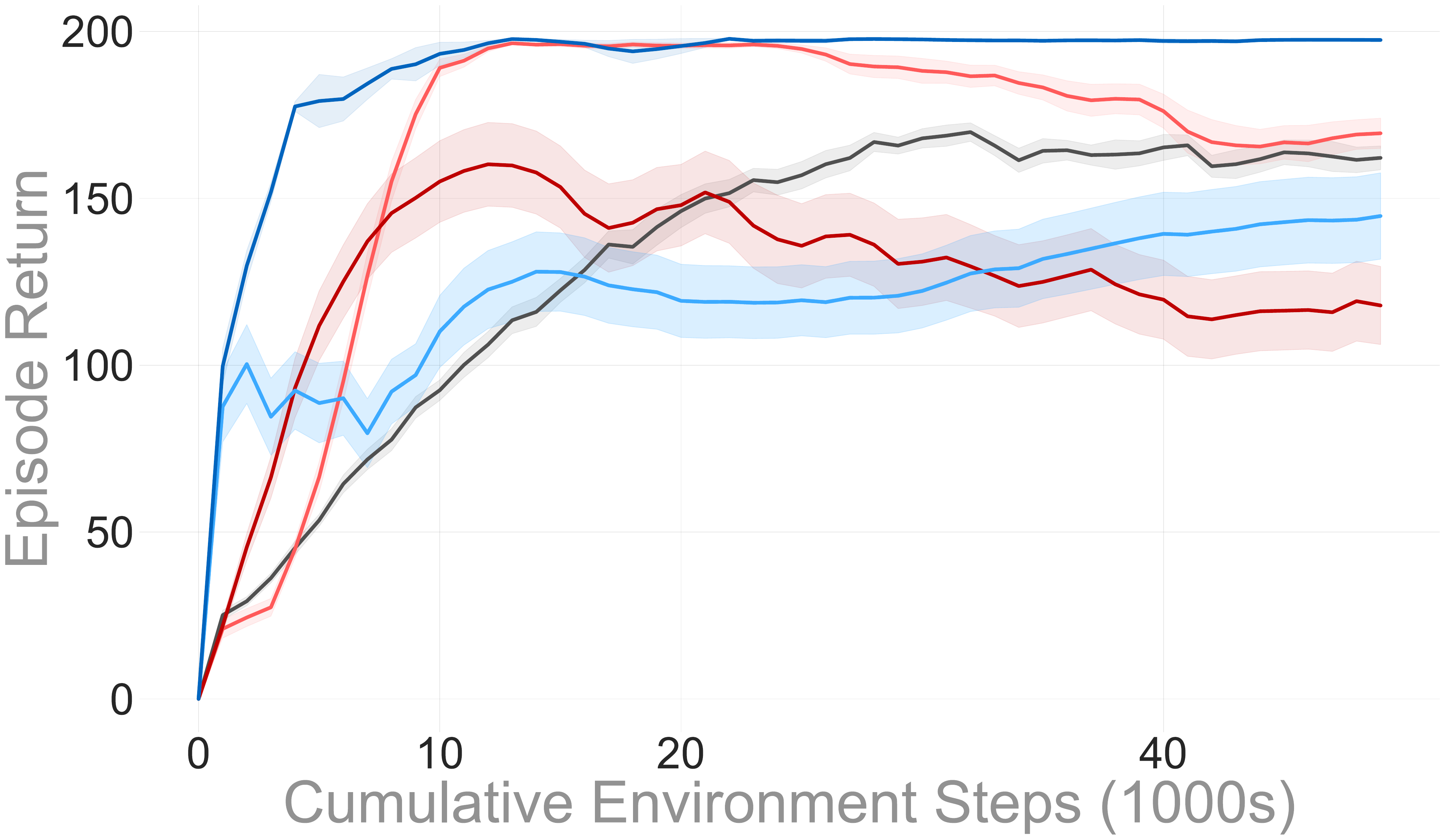}}
	\subfloat[\emph{Catcher}]{\includegraphics[width=0.5\columnwidth]{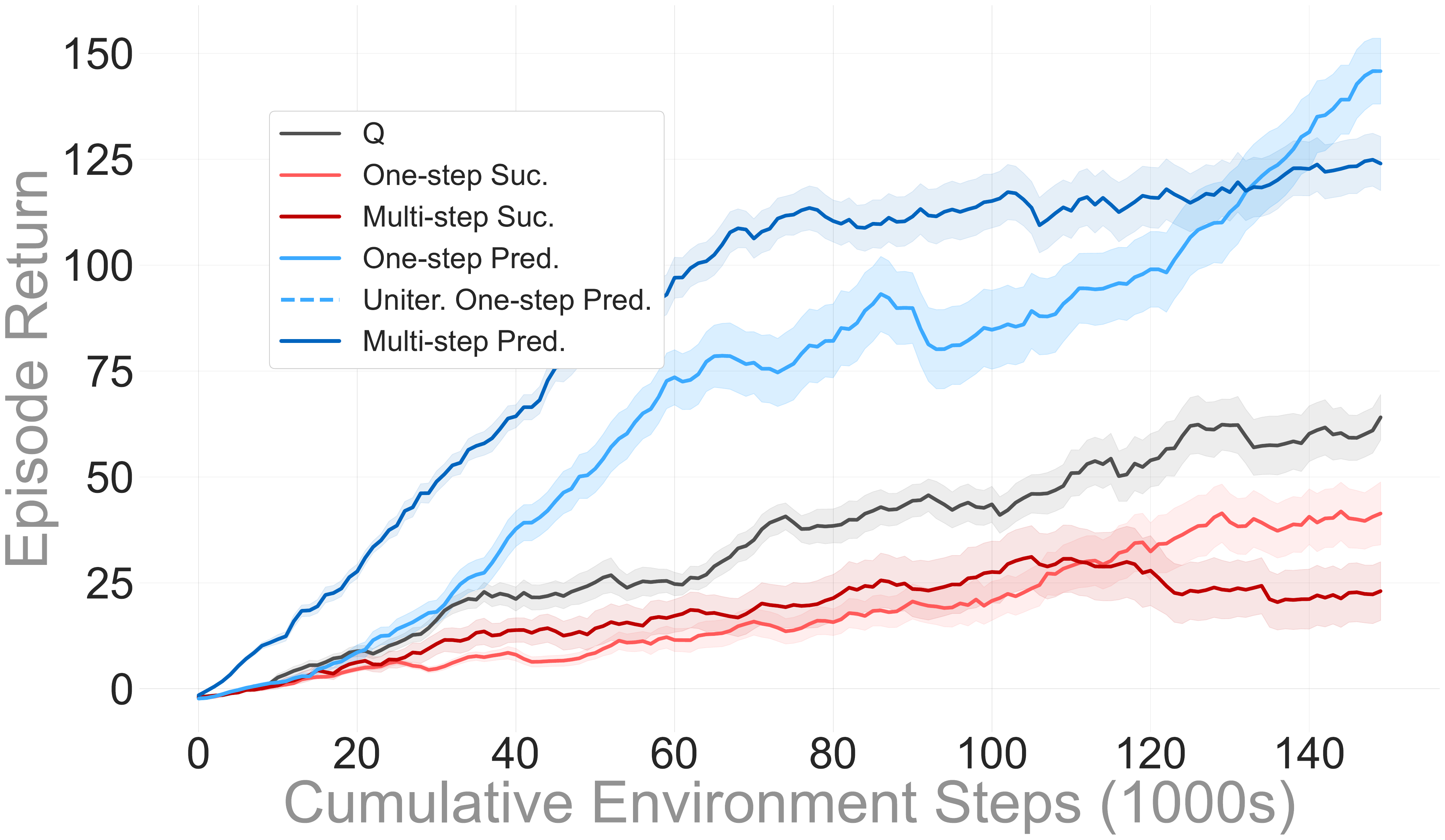}}\\
	\subfloat[\emph{Puddleworld}]{\includegraphics[width=0.5\columnwidth]{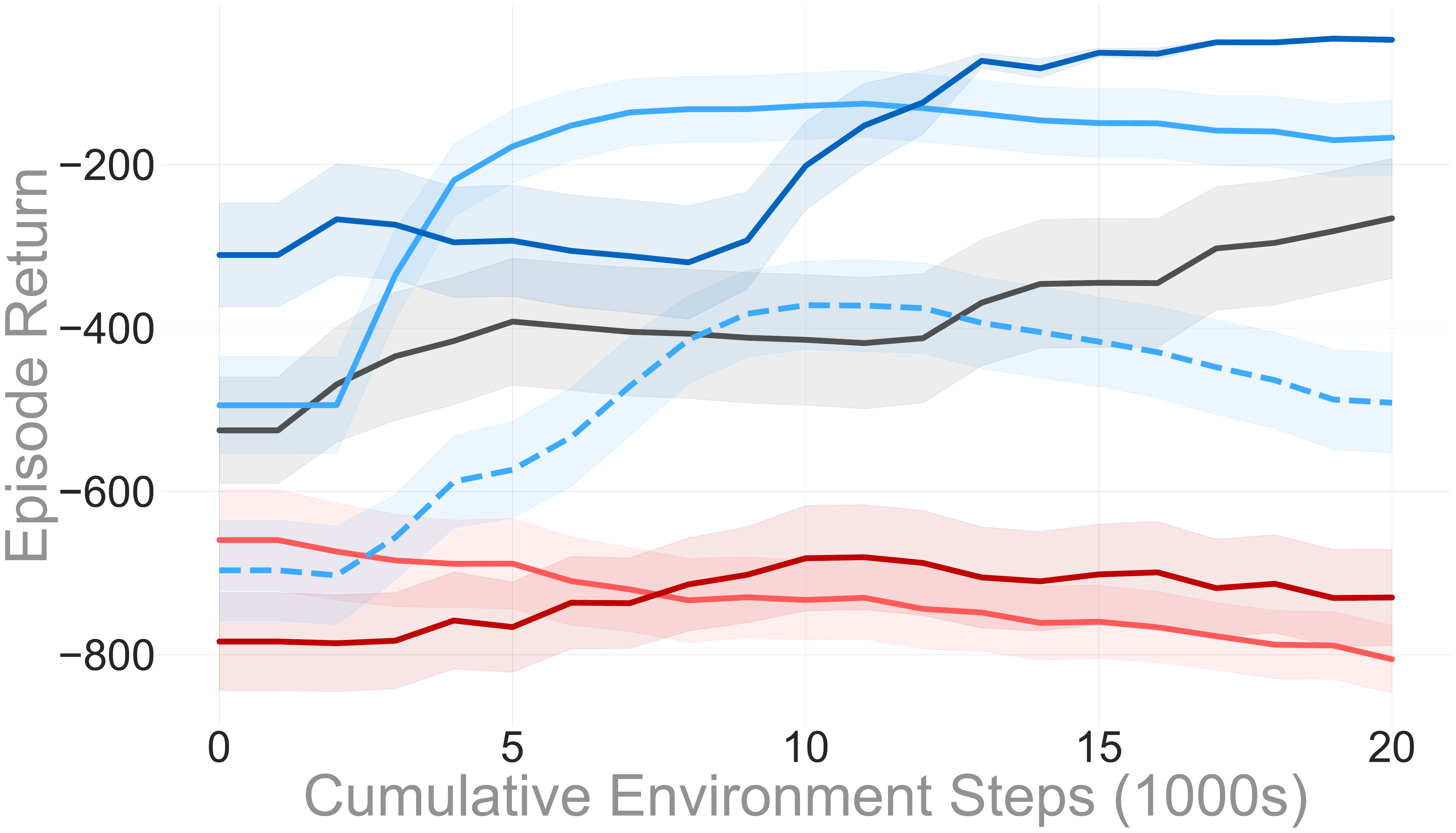}}
	
	\caption{Learning curves for the Dyna variants while updating the environment model \textbf{online} with a conservative screening approach. All curves are averaged over 30 runs, with shaded regions corresponding to standard errors. Multi-step Predecessor performs better in both Catcher and Puddleworld. If an environment model is less prone to HVH, the performance of all variants gets closer to each other, e.g. Cartpole. In online learning, since the model is updated by seeing more samples, the performance of One-step methods might end up being higher up since they perform more planning updates in the background.}
	\label{Figure:Section6/LearningCurvesOnlineconservative}
\end{figure}

\section{Learning Curves on Benchmarks for Conservative Screening Approach} \label{appendix B}

Figures \ref{Figure:Section6/LearningCurvesOfflineconservative} and \ref{Figure:Section6/LearningCurvesOnlineconservative} show the learning curves of offline and online learning when we apply conservative screening on three benchmarks. The results demonstrates the superiority of multi-step backwards in all settings except Catcher, where Uinterated predecessor surpasses the multi-step backwards at the later learning steps since they get to perform more updates while also benefit from the HVH avoidance property of uniterated backwards.

\vskip 0.2in
\bibliography{paper}
\bibliographystyle{theapa}

\end{document}